\documentclass[acmlarge]{acmart}

\AtBeginDocument{%
  \providecommand\BibTeX{{%
    \normalfont B\kern-0.5em{\scshape i\kern-0.25em b}\kern-0.8em\TeX}}}

\newif\ifacmartsty
\acmartstytrue
\newif\ifcameraready
\camerareadytrue

\usepackage[T1]{fontenc}
\usepackage[utf8]{inputenc}

\usepackage{upquote}

\usepackage{microtype}
\UseMicrotypeSet[protrusion]{basicmath} 

\usepackage{hyperref}
\hypersetup{
  colorlinks   = true, 
  urlcolor     = blue, 
  linkcolor    = red, 
  citecolor   = blue 
}
\usepackage{graphicx,grffile}

\usepackage{mathrsfs}
\usepackage{gensymb}
\usepackage{multirow}
\usepackage{xspace}
\usepackage{tabularx}
\usepackage{ragged2e}
\usepackage{booktabs}
\usepackage{paralist}
\usepackage[american]{babel}
\usepackage[shortlabels]{enumitem}
\usepackage{courier,color,wrapfig}
\usepackage{xspace}
\usepackage{balance}
\usepackage{enumitem}
\usepackage{epstopdf}
\usepackage{multirow}
\usepackage{booktabs}
\usepackage{url}
\usepackage{pifont}
\usepackage{subcaption}
\usepackage{tikz}
\usepackage{comment}
\usepackage{mathtools}
\usepackage{amsmath}
\usepackage[normalem]{ulem}
\usepackage{xkeyval}
\usepackage{algorithm}
\usepackage[noend]{algpseudocode}

\usepackage[textsize=small,disable]{todonotes}

\usepackage{etoolbox}
\usepackage{ifthen}
\usepackage{sepfootnotes}

\usepackage{caption}
\usepackage[short]{optidef} 

\makeatletter
\def\maxwidth{\ifdim\Gin@nat@width>\linewidth\linewidth\else\Gin@nat@width\fi}
\def\maxheight{\ifdim\Gin@nat@height>\textheight\textheight\else\Gin@nat@height\fi}
\makeatother
\setkeys{Gin}{width=\maxwidth,height=\maxheight,keepaspectratio}
\makeatletter
\g@addto@macro{\UrlBreaks}{\UrlOrds}
\makeatother

\makeatletter
\def\algbackskip{\hskip-\ALG@thistlm}
\makeatother

\ifacmartsty
\makeatletter
\let\origsection\section
\let\origsubsection\subsection

\renewcommand\section{\@ifstar{\starsection}{\nostarsection}}
\renewcommand\subsection{\@ifstar{\starsubsection}{\nostarsubsection}}

\newcommand\sectionprelude{\vspace{0.25ex}}
\newcommand\sectionpostlude{\vspace{0.25ex}}
\newcommand\subsectionprelude{\vspace{0.5ex}}
\newcommand\subsectionpostlude{\vspace{0.5ex}}

\newcommand\nostarsection[1]{\sectionprelude\origsection{#1}\sectionpostlude}
\newcommand\starsection[1]{\sectionprelude\origsection*{#1}\sectionpostlude}

\newcommand\nostarsubsection[1]{\subsectionprelude\origsubsection{#1}\subsectionpostlude}
\newcommand\starsubsection[1]{\subsectionprelude\origsubsection*{#1}\subsectionpostlude}

\makeatother
\fi

\apptocmd\normalsize{%
\abovedisplayskip=3pt
\abovedisplayshortskip=3pt
\belowdisplayskip=3pt
\belowdisplayshortskip=3pt
}{}{}

\ifacmartsty

\setcopyright{acmlicensed}
\acmJournal{IMWUT}

\fi

\newcommand{\sysname}{\textsc{AeroTraj}\xspace}
\newcommand{\system}{\textsc{AeroTraj}\xspace}

\newcommand{\etc}{\emph{etc.}\xspace}
\newcommand{\ie}{\emph{i.e.,}\xspace}
\newcommand{\eg}{\emph{e.g.,}\xspace}

\newcommand{\secref}[1]{\S\ref{#1}}
\newcommand{\figref}[1]{Fig.~\ref{#1}}
\newcommand{\tabref}[1]{Table~\ref{#1}}
\newcommand{\algoref}[1]{Algorithm~\ref{#1}}
\newcommand{\eqnref}[1]{Equation~\ref{#1}}

\newif\ifhighlightrevisions
\highlightrevisionsfalse

\newif\ifhighlightminorrevisions
\highlightminorrevisionsfalse

\ifhighlightrevisions

\newcommand{\add}[1]{\textcolor{blue}{#1}}

\else

\newcommand{\add}[1]{\textcolor{black}{#1}}

\fi

\ifhighlightminorrevisions
\newcommand{\mradd}[1]{\textcolor{blue}{#1}}
\else
\newcommand{\mradd}[1]{{#1}}
\fi

\newcommand{\fawad}[1]{\todo[author=Fawad,color=orange,inline]{#1}}

\newcommand{\raj}[1]{\todo[author=Raj,color=cyan!25,inline]{#1}}

\begin{document}
\title{\sysname: Trajectory Planning for Fast, and Accurate 3D Reconstruction Using a Drone-based LiDAR}

\author{Fawad Ahmad}
\email{fawad@cs.rit.edu}
\orcid{0000-0003-4182-229X}
\affiliation{%
  \institution{Rochester Institute of Technology}
  \country{USA}
}

\author{Christina Suyong Shin}
\email{cshin956@usc.edu}
\orcid{0000-0001-5653-5571}
\affiliation{%
  \institution{University of Southern California}
  \country{USA}
}

\author{Rajrup Ghosh}
\email{rajrupgh@usc.edu}
\orcid{0000-0002-9797-2417}
\affiliation{%
  \institution{University of Southern California}
  \country{USA}
}

\author{John D'Ambrosio}
\email{jfdambro@usc.edu}
\orcid{0009-0001-1446-8427}
\affiliation{%
  \institution{University of Southern California}
  \country{USA}
}

\author{Eugene Chai}
\email{eugene.chai@nokia-bell-labs.com}
\orcid{0009-0003-1505-7368}
\affiliation{%
  \institution{Nokia Bell Labs}
  \country{USA}
}

\author{Karthikeyan Sundaresan}
\email{karthik@ece.gatech.edu}
\orcid{0000-0002-8089-4264}
\affiliation{%
  \institution{Georgia Institute of Technology}
  \country{USA}
}

\author{Ramesh Govindan}
\email{ramesh@usc.edu}
\orcid{0000-0001-8311-8853}
\affiliation{%
  \institution{University of Southern California}
  \country{USA}
}

\renewcommand{\shortauthors}{Ahmad et al.}

\begin{abstract}
  This paper presents \sysname, a system that enables fast, accurate, and automated reconstruction of 3D models of large buildings using a drone-mounted LiDAR.
  LiDAR point clouds can be used directly to assemble 3D models if their positions are accurately determined.
  \sysname uses SLAM for this, but must ensure complete and accurate reconstruction while minimizing drone battery usage. Doing this requires balancing competing constraints: drone speed, height, and orientation. \sysname exploits building geometry in designing an optimal trajectory that incorporates these constraints. Even with an optimal trajectory, SLAM's position error can drift over time, so \sysname tracks drift in-flight by offloading computations to the cloud and invokes a re-calibration procedure to minimize error. \sysname can reconstruct large structures with centimeter-level accuracy and \mradd{with an average end-to-end latency below 250~ms}, significantly outperforming the state of the art. 
\end{abstract}



\setcopyright{acmlicensed}
\acmJournal{IMWUT}
\acmYear{2023} \acmVolume{7} \acmNumber{3} \acmArticle{83} \acmMonth{9} \acmPrice{15.00}\acmDOI{10.1145/3610911}

\ccsdesc[300]{Human-centered computing~Ubiquitous and mobile computing systems and tools}

\keywords{3D Reconstruction, Mapping, Trajectory Planning, Localization}

\maketitle

\section{Introduction}
\label{sec:intro}

Drones have developed to the point where they can be equipped with on-board compute, cellular radios (LTE) and sophisticated sensors like stereo cameras, and LiDARs\add{,} \etc
This has spurred interest in drone-based \textit{3D reconstruction} of buildings and other large structures for construction site monitoring~\cite{equinoxdrones}, damage assessment~\cite{drone_disaster_relief_1, drone_disaster_relief_2, drone_disaster_relief_3, company_austrian_airline}, and documenting repair and retrofit tasks~\cite{company_drone_deploy}. In these applications, a drone captures imagery outside a building, and uses it to generate a \textit{3D model} of the building.


The term \textit{3D model} covers a range of geometric representations of the surfaces of objects, from coarse-grained approximations (cylinders, cubes, intersection of planes), to more fine-grained representations such as \textit{meshes} (small-scale surface tessellations that capture structural variations). In this paper, we seek to extract an even finer-grained \textit{point-cloud} of a large structure (\eg a building) which consists of dense \textit{points} on the surface of the structure. Each point has an associated 3D position, along with other attributes (depending on the sensor used to generate the point-cloud). A point-cloud based 3D model can generate all other representations.
Today, to build a 3D model of a building, one can fly a drone around the building in a given trajectory to collect 2D images then use \textit{photogrammetry} which infers a 3D model from a sequence of 2D images~\cite{7139681, 7989530, company_hover}. Photogrammetry is \textit{computationally intensive} and can require multiple iterations, each requiring human intervention (\secref{sec:background_motivation}).

Unlike photogrammetry, LiDAR-based reconstruction can more directly infer 3D models, because LiDARs directly provide depth information (unlike cameras). A LiDAR sensor measures distances to surfaces using lasers mounted on a mechanical rotating platform. The number of lasers determines the resolution of the LiDAR. With each revolution of the lasers, the LiDAR returns a point cloud, or a \textit{LiDAR frame}. The point cloud is a set of 3D data points, each corresponding to a distance measurement of a particular position of the surrounding environment from the LiDAR.


To obtain a high quality reconstruction of a structure such as a building, one can mount a LiDAR on a drone, scan the building by flying the drone around it, and \textit{merge} points clouds captured from different locations around the building. Consider point clouds $p$ and $p'$. A point $x$ on the surface of the building may appear both in $p$ and $p'$. However, since the drone has moved, this point $x$ appears at different positions (relative to the LiDAR) in the two point clouds. If we can position the drone accurately at each instant, we can precisely transform both point clouds to the same coordinate frame of reference. Then, the union of points in $p$ and $p'$ constitutes part of the 3D model of the building. To obtain the complete model, the drone must scan the entire building using a \textit{trajectory} of minimal flight duration to minimize drone battery usage.

GPS is inadequate to position the drone accurately (\secref{sec:background_motivation}), but LiDAR SLAM (Simultaneous Localization and Mapping~\cite{slam_part1, slam_part2}) is a potential candidate. LiDAR SLAM can provide centimeter-level positioning for autonomous vehicles.
However, in our setting, off-the-shelf LiDAR SLAM does not work well (\secref{sec:background_motivation}), because we use the LiDAR for positioning \textit{and} reconstruction \textit{and} must respect drone battery constraints. These goals are mutually in conflict (\secref{sec:design}) because SLAM determines pose transformations between successive point clouds, and accurate transformations need a significant number of \textit{overlapped} points between point clouds. To ensure overlap:
\begin{itemize}[nosep,leftmargin=*]
\item The drone can fly slowly, but this results in increased flight duration. 
\item It can fly further from the building surface. This way, the LiDAR can capture more of the surface, and the drone can fly faster. However, LiDAR point resolution decreases with distance, resulting in a poor reconstruction.
\item Even if we can address these competing constraints, SLAM's position estimates can drift over time. If not corrected, drift can adversely impact reconstruction quality. 
\end{itemize}

In this paper, we describe the design, implementation and evaluation of \sysname, which addresses these challenges using two insights: it (a) exploits building geometry to design a scan trajectory that enables high quality reconstruction while minimizing flight time, and (b) detects and corrects SLAM drift in-flight. The paper makes the following contributions (\secref{sec:design}):
\mradd{
\begin{itemize}[nosep,leftmargin=*]
\item It presents a novel optimization formulation that generates minimum-flight-time \textit{model collection} trajectories for a large class of buildings while respecting speed, height, and LiDAR orientation and resolution constraints.
\item It proposes a cloud-offload architecture for automated in-flight drift estimation, so drone flights can be \textit{re-calibrated} when drift becomes too high.
\item It uses a fast and efficient \textit{reconnaissance} flight design to estimate building geometry parameters for trajectory generation.
\end{itemize}
}

\mradd{As a by-product of this design, \sysname can construct models on-the-fly: the 3D model is available within hundreds of milliseconds of flight completion.} Experiments (\secref{sec:eval}) with a complete \sysname implementation on real-world flights and a photorealistic drone simulator (AirSim~\cite{airsim}) demonstrate that \sysname can reconstruct 3D models to within 10~cm accuracy, for a variety of building shapes and sizes. Moreover, \sysname is \textit{faster}, and \textit{more accurate} than existing state-of-the-art approaches, photogrammetry~\cite{colmap}, and LiDAR-based 3D reconstruction~\cite{zhang2014loam, Cartographer}.
\section{Background and Motivation}
\label{sec:background_motivation}

In this section, we describe metrics that capture the quality of 3D reconstruction, and results from experiments that motivate \sysname.

\subsection{Measures of 3D Model Quality}

Prior work on 3D reconstruction~\cite{reconstructionmetrics} has proposed two metrics for reconstruction quality: \textit{accuracy} and \textit{completeness}.
Consider a 3D model $M$ and a corresponding ground-truth $M_g$. 
Accuracy describes how \textit{closely} the 3D model $M$ resembles the ground truth $M_g$. 
Accuracy is the root mean square error (RMSE) of the distance from each point in $M$ to the nearest point in $M_g$. 
Completeness describes \textit{what extent} of the ground truth $M_g$ the 3D model $M$ captures.
Completeness is the RMSE of the distance from each point in $M_g$ to the nearest point in $M$. 

\begin{figure}[t]
\captionsetup[subfigure]{labelfont=rm}
\centering
    \begin{subfigure}{0.33\columnwidth}
      \includegraphics{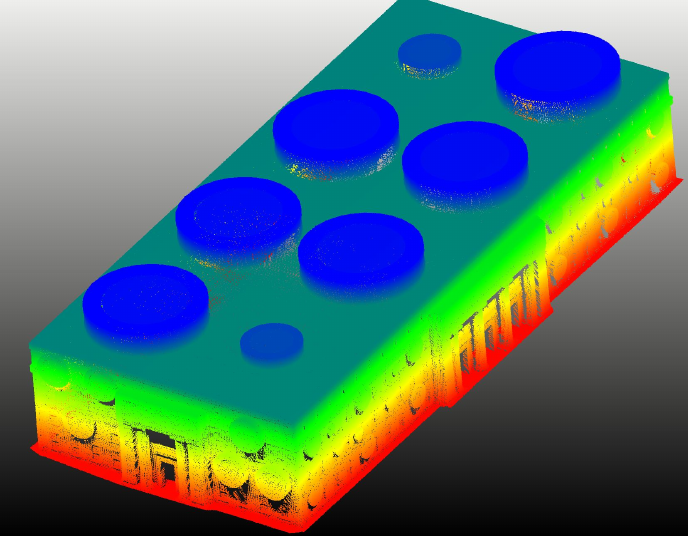}
      \caption{\mradd{Ground truth 3D model.}}
      \label{fig:building_gt}
    \end{subfigure}
    \begin{subfigure}{0.33\columnwidth}
      \includegraphics{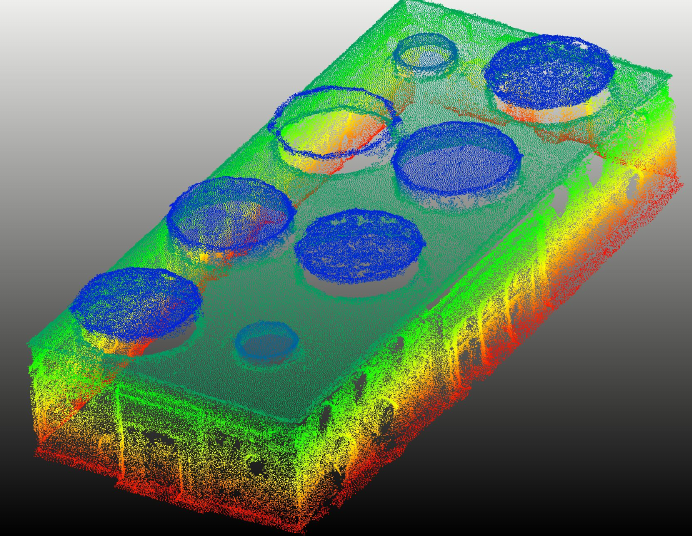}
      \caption{\mradd{MVS/photogrammetry-dervied 3D model.}}
      \label{fig:building_mvs}
    \end{subfigure}
    \caption{Compared to a ground truth 3D model (a), an MVS/photogrammetry-based 3D reconstruction (b) of a building can be \textit{incomplete} \ie it contains holes and missing regions.}
    \label{fig:mvs}
\end{figure}

A 10~cm accuracy means that every point on the 3D model $M$ is displaced by 10~cm, on average, from its ground truth position in $M_g$.
A 50~cm completeness means that, on average, for every point on the ground truth $M_g$, the nearest point on the 3D model $M$ is 50~cm away.
If both accuracy and completeness are zero, $M$ perfectly matches $M_g$.
So, for both metrics, \textit{lower is better}. 
These two metrics are roughly analogous to precision (accuracy) and recall (completeness).
\add{\figref{fig:building_mvs} shows an \textit{accurate but incomplete} 3D reconstruction.}

{\small
\begin{table}
  \centering
    \caption{\add{Photogrammetry can be slow, inaccurate and incomplete. The small building is 50m~x~50m~x~20m (length x breadth x height), the large building is 100m~x~50m~x~20m.} 
  }

  \vspace{0.1cm}
  \begin{tabular}{{c  c  c  c}} \hline \hline
    \textbf{Building} & \textbf{Accuracy (m)} & \textbf{Completeness (m)} & \textbf{Time (s)} \\ \hline
    Small  & 0.11 & 0.80 & 23084 \\ 
    Large  & 0.16 & 0.75 & 31600 \\ \hline
\vspace{-0.0cm} &
    
  \end{tabular}
  \label{tab:mvs_motiv}
\end{table}
}

\subsection{Photogrammetry} 
Photogrammetry, a technique to construct 3D models, uses multi-view stereo~\cite{furukawa15:_multi_view_stereo} (MVS). MVS infers the scene's 3D structure, and the depth of every pixel in all the images. Then, it fuses these into a dense 3D reconstruction. MVS 3D reconstructions are generally accurate, but may contain missing regions or holes (\figref{fig:building_mvs}) when it cannot confidently estimate pixel depths. Moreover, MVS is also compute-intensive.

\add{In our setting, to build a model using MVS, an operator would fly a drone to collect 2D images. Then, after landing the drone, she would upload the images to a cloud service to run MVS.
}
If the model \textit{contains missing regions} (detected by visual inspection), the operator must adjust the trajectory and fly the drone again~\cite{mvslessons, photogrammetryissues}. 
It can take multiple iterations to get the desired 3D model. 

To quantify these shortcomings, we conducted an experiment (methodology described in~\secref{sec:eval}) where we used an open source implementation of MVS (ColMap~\cite{colmap}) to reconstruct 3D models of two buildings. 
For each building, we tried several trajectories, and selected the one that provided the highest quality model. 
MVS accuracy is slightly higher than 10~cm (\tabref{tab:mvs_motiv}); this is desirable. 
Completeness, however, is 75~cm or more.
\figref{fig:building_mvs} illustrates incomplete regions in the MVS reconstruction.
\sysname achieves 5 to 9~cm completeness (\secref{sec:eval}), almost an order of magnitude better. 
\add{Finally, MVS requires nearly 7–9 \textit{hours} after flight completion to reconstruct the model. \sysname assembles the model on-the-fly so the model is ready immediately after flight completion.}

\add{
In theory, photogrammetry could detect, in-flight, when the model is incomplete and adapt the trajectory accordingly. Detecting model incompleteness is challenging without running MVS on the entire set of images~\cite{mostegel2016uav, 7422384}. To reduce the number of iterations, companies hire drone pilots or photogrammetrists~\cite{mvslessons, photogrammetryissues} who design trajectories based on their experience and rules of thumb~\cite{colmap, dd_guidelines}. Even so, it is very challenging to ensure complete and high-quality reconstructions without iteration~\cite{mostegel2016uav}. 
}


\subsection{GPS}

\begin{figure}[t]
\captionsetup[subfigure]{labelfont=rm}
\centering
    \begin{subfigure}{0.3\linewidth}
      \includegraphics{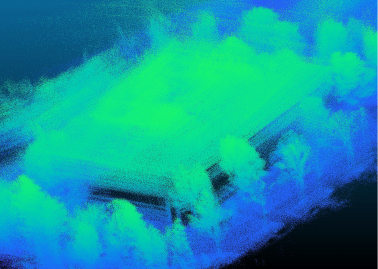}
      \caption{\mradd{GPS.}}
      \label{fig:model_comp_gps}
    \end{subfigure}
    \begin{subfigure}{0.3\linewidth}
      \includegraphics{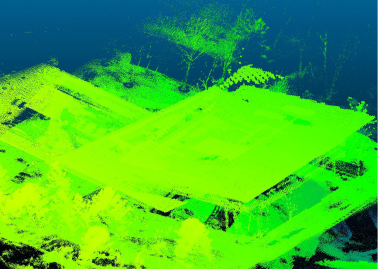}
      \caption{\mradd{SLAM.}}
      \label{fig:model_comp_slam}
    \end{subfigure}
    \begin{subfigure}{0.2825\linewidth}
      \includegraphics{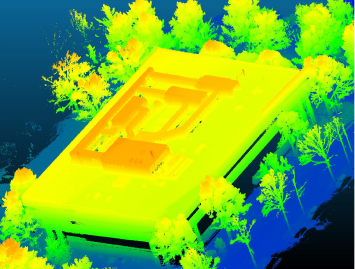}
      \caption{\mradd{\sysname}.}
      \label{fig:model_comp_aerotraj}
    \end{subfigure}
    \caption{Three models of a large complex on our campus: GPS (\mradd{a}), SLAM (\mradd{b}), and \sysname (\mradd{c}). All models use the same height ramp function to color-code the z-axis, with blue representing the lowest points and orange representing the highest. The \sysname model has distinct features and trees, with clear and crisp coloring, indicating a good reconstruction. In contrast, the GPS and SLAM models have diffused colors and lack visible features, suggesting poor and noisy reconstruction.}
    \label{fig:slam_vs_gps_3d_model}
\end{figure}

On-board GPS receivers can be used to position point clouds. Unfortunately, GPS errors can range up-to several meters~\cite{gnome}.
\figref{fig:model_comp_gps} shows a 3D model assembled using a drone flight over a commercial building that uses GPS for positioning; the building's outline is fuzzy, as are the contours of the trees surrounding the building. \tabref{tab:slam_gps_reconstruction} shows that the accuracy and completeness for the GPS reconstructed model are 1.6~m and 0.53~m, respectively, both of which are undesirable. \figref{fig:model_comp_aerotraj} shows a 3D reconstruction using techniques proposed in this paper, which does not use GPS. 


{\small
\begin{table}
  \caption{Relative to \sysname, alternative reconstruction approaches for drone-based LiDAR reconstruction give poor quality 3D models.}

  \begin{tabular}{{c c c c}} \hline \hline
    \textbf{Approach} & \textbf{Accuracy (m)} & \textbf{Completeness (m)} & \add{\textbf{Flight time (s)}} \\ \hline
    GPS  & 1.60 & 0.53 & \add{944} \\ 
    SLAM  & 2.30 & 1.30 & \add{944} \\ 
    \sysname & \textbf{0.13} & \textbf{0.09} & \textbf{\add{750}} \\ \hline
\vspace{-0.0cm} &
    
  \end{tabular}
  \label{tab:slam_gps_reconstruction}
\end{table}
}

High-precision GNSS/RTK receivers can provide more accurate positioning but require additional infrastructure, are costly, and can perform poorly (tens of meters of error~\cite{Carloc}) in urban environments. Prior work~\cite{uav_lidar_1,uav_lidar_2} has used such receivers in \textit{remote sensing} for \textit{offline} reconstruction, using specialized unmanned aerial vehicles (UAVs) with long-range LiDAR sensors.
In contrast, we consider solutions that employ off-the-shelf technologies: drones, and commodity GPS and LiDAR for online reconstruction.

\subsection{SLAM}

In this paper, we use SLAM to ensure accurate drone positioning. LiDAR-based SLAM algorithms continuously align 3D point clouds using scan~\cite{Cartographer}, and feature matching~\cite{zhang2014loam,liosam} techniques to precisely estimate the pose (position, and orientation) of the LiDAR. These algorithms can obtain centimeter-level accuracy for LiDARs mounted on autonomous vehicles.

\begin{figure}[t]
\centering
\begin{minipage}{0.65\linewidth}
 \centering\includegraphics[width=0.80\textwidth]{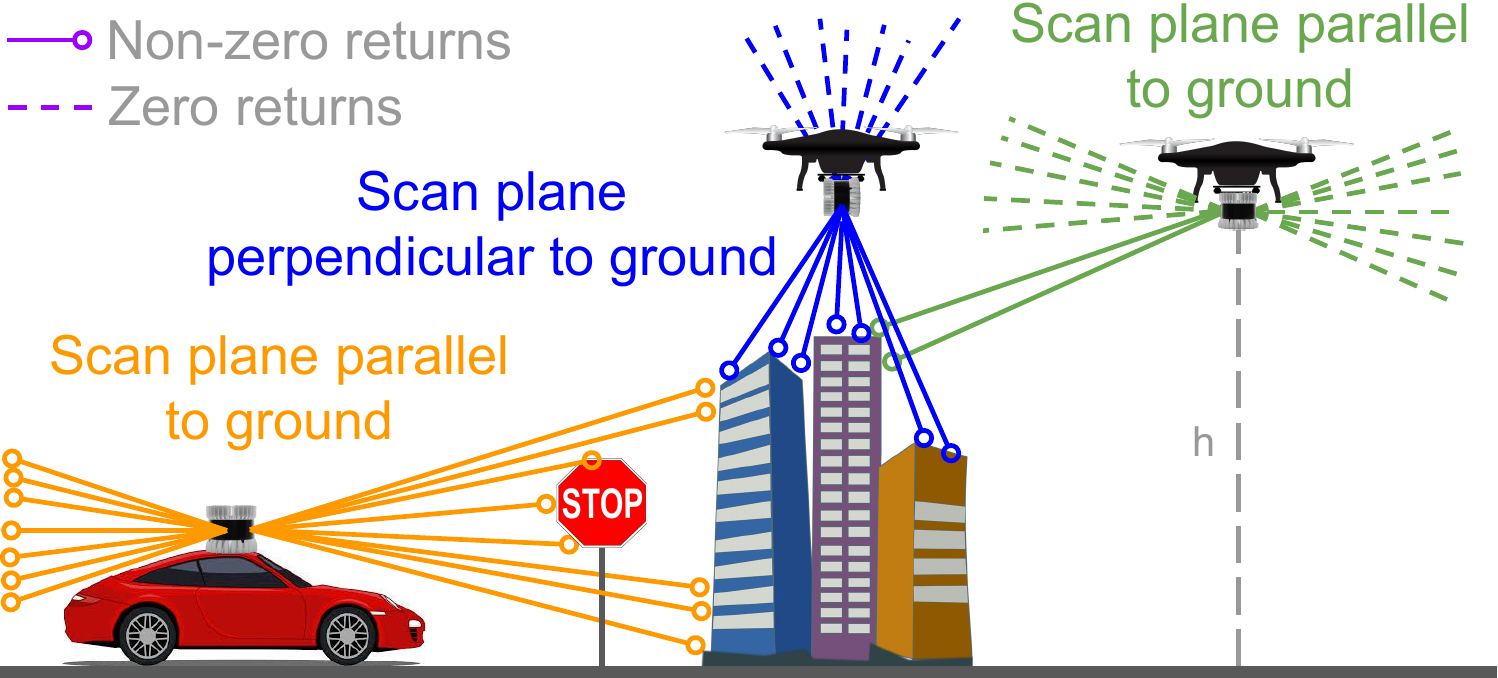}
 \caption{\mradd{LiDAR returns when mounted on a vehicle (orange lines) and a drone platform (blue, and green lines). Solid lines represent the laser beams which hit other objects (non-zero returns), whereas the dotted lines represent ones that did not (zero returns).}}
 \label{fig:slam_challenge}
\end{minipage}
\hfill
\begin{minipage}{0.3\linewidth}
 \includegraphics{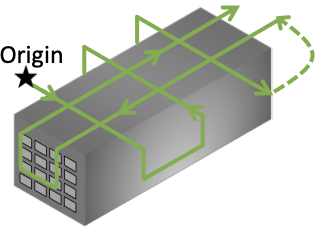}
 \caption{Rectilinear trajectory for SLAM-based reconstruction.}
 \label{fig:rectflight}
\end{minipage}
\end{figure}

\add{
On a drone, however, un-modified SLAM cannot obtain similar accuracy. To understand why, consider \figref{fig:slam_challenge}. On a vehicle, the LiDAR scan plane (the plane formed by the laser beams in the center) is parallel to the ground. This way, the LiDAR receives numerous returns \mradd{(solid orange lines)} from objects in the surrounding \eg the road, traffic signs, and buildings, \etc This results in dense point clouds, better alignment and hence accurate SLAM positioning. On a drone at height $h$, a similar orientation  \mradd{(green lines in \figref{fig:slam_challenge})} results in significantly fewer returns (if any).
When the scan plane is perpendicular to the ground \mradd{(blue lines in \figref{fig:slam_challenge})}, the LiDAR receives more returns but only about 9\% of the returns from a vehicle-mounted LiDAR. With sparse returns, SLAM cannot precisely align consecutive point clouds and hence results in imprecise and inaccurate positioning. Thus, LiDAR orientation with respect to the building surface is an important factor in determining accuracy; we discuss other factors in \secref{sec:design}.
}


To quantify this, we used a rectilinear flight (\figref{fig:rectflight}) to build a 3D model of a large building using the Google Cartographer LiDAR SLAM~\cite{Cartographer} implementation.
The resulting 3D model (\figref{fig:model_comp_slam}) has 2.3~m accuracy, 1.3~m completeness (\tabref{tab:slam_gps_reconstruction}), and is visually worse than \sysname-based reconstruction (\figref{fig:model_comp_aerotraj}).
SLAM cannot align successive point clouds and hence results in imprecise pose estimates.

\sysname is significantly more accurate and complete than un-modified SLAM. It uses SLAM as a starting point in its design. However, for \sysname to detect SLAM drift (\secref{sec:intro}) in-flight, it must run SLAM as the drone flies. Because of payload restrictions, on-board compute on drones may be insufficient to run SLAM.
A DJI M600Pro (which we use in this paper) hexacopter has a maximum payload weight of 5~kg. 
On this, we mounted an LTE radio, an Ouster OS1-64 LiDAR and a Jetson TX2 board.
This compute capability is far from sufficient to run LiDAR SLAM at full-frame rate\footnote{With a 64-beam LiDAR, SLAM processes up to 480 Mbps of 3D \mradd{data.}}. On the TX2, we ran two popular, representative LiDAR-SLAM algorithms, Cartographer~\cite{Cartographer} and LOAM~\cite{zhang2014loam}.
Cartographer, a scan matching algorithm, can only process 2 frames per second (LiDARs generate 10–20 frames per second) and LOAM, a feature-matching algorithm, can process 5 frames per second (\figref{tab:lidar_slam_compute}).  LOAM's reconstruction quality is unacceptable. Cartographer produces good reconstructions, but its frame rate is too slow to detect and correct drift in-flight using on-board compute (\tabref{tab:lidar_slam_compute} shows results for a short flight in which drift is not a factor).
 
\begin{table}[t]
  \small
    \caption{LiDAR SLAM on Jetson TX2. For both accuracy and completeness, lower is better.}
  {
  \vspace{0.1cm}
  
  \centering
  \begin{tabular}{c  c  c} \hline \hline 
    \textbf{Metric} & \textbf{Cartographer~\cite{Cartographer}} & \textbf{LOAM~\cite{zhang2014loam}} \\ \hline
    Frame Rate (fps) & 2 & 5 \\ 
    Accuracy (m) & 0.21 & 5.75 \\ 
    Completeness (m) & 0.09 & 2.74 \\ \hline
  \end{tabular}
  }

  \label{tab:lidar_slam_compute}
\end{table}


These illustrative results motivate the design of \sysname, which layers trajectory generation and drift correction on top of SLAM to achieve centimeter-level accuracy for LiDAR-on-drone based 3D reconstruction.

\section{\sysname Design}
\label{sec:design}

\begin{table}[t]
    \centering
        \caption{\sysname can reconstruct more than 99\% commercial and residential buildings in four representative cities.}
    \small
    {
  \vspace{0.1cm}
    
    \begin{tabular}{c  c  c  c  c  c}
    \hline
    \textbf{Roof type} & \textbf{\sysname support} & 
    \textbf{Witten} & 
    \textbf{Ann Arbor} &
    \textbf{Manhattan} & 
    \textbf{Farmingham} 
     \\ \hline \hline
    Polygonal & \checkmark& 709  & 1686 & 75 & 5      \\ 
    Gabled    & \checkmark& 572  & 7    & 96 & 4047 \\ 
    Hipped    & \checkmark& 1027 & 3    & 23 & 538  \\ 
    Pyramidal & \checkmark& 110  & 0    & 0  & 0  \\ 
    Skillion  & \checkmark& 189  & 0    & 2  & 0 \\ 
    Gambrel   & \checkmark& 0    & 0    & 0  & 143 \\ 
    Others    & \checkmark& 130  & 772  & 14 & 0 \\ 
    Dome/antenna & $\times$ & 3      & 20        & 0         & 1         \\ 
    \multicolumn{2}{c}{\begin{tabular}[c]{@{}c@{}}Percentage of buildings \\ \sysname can reconstruct \end{tabular}}
    & 99.8\% & 99.1\% & 100.0\% & 99.9\%  \\   \hline
    \vspace{0cm} &
    \end{tabular}
    }

    \label{tab:roof_types}
    \end{table}


\begin{figure}
\captionsetup[subfigure]{labelfont=rm}
\centering
    \begin{subfigure}{0.2\linewidth}
      \includegraphics{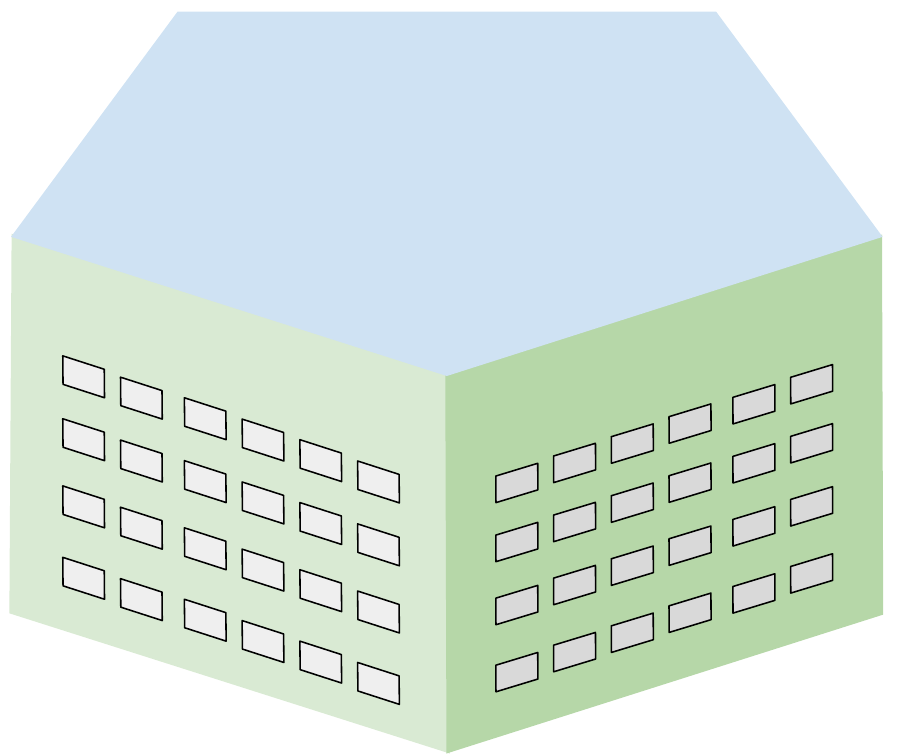}
      \caption{Pentagon.}
      \label{fig:pentagon_building}
    \end{subfigure}
    \begin{subfigure}{0.2\linewidth}
      \includegraphics{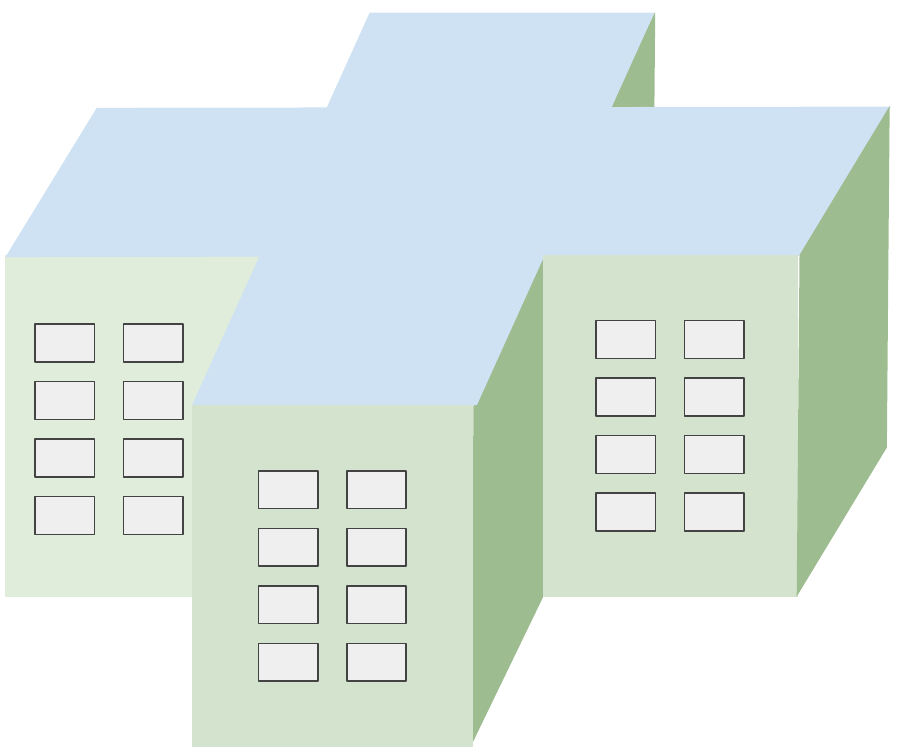}
      \caption{Plus.}
      \label{fig:plus_building}
    \end{subfigure}
    \begin{subfigure}{0.2\linewidth}
      \includegraphics{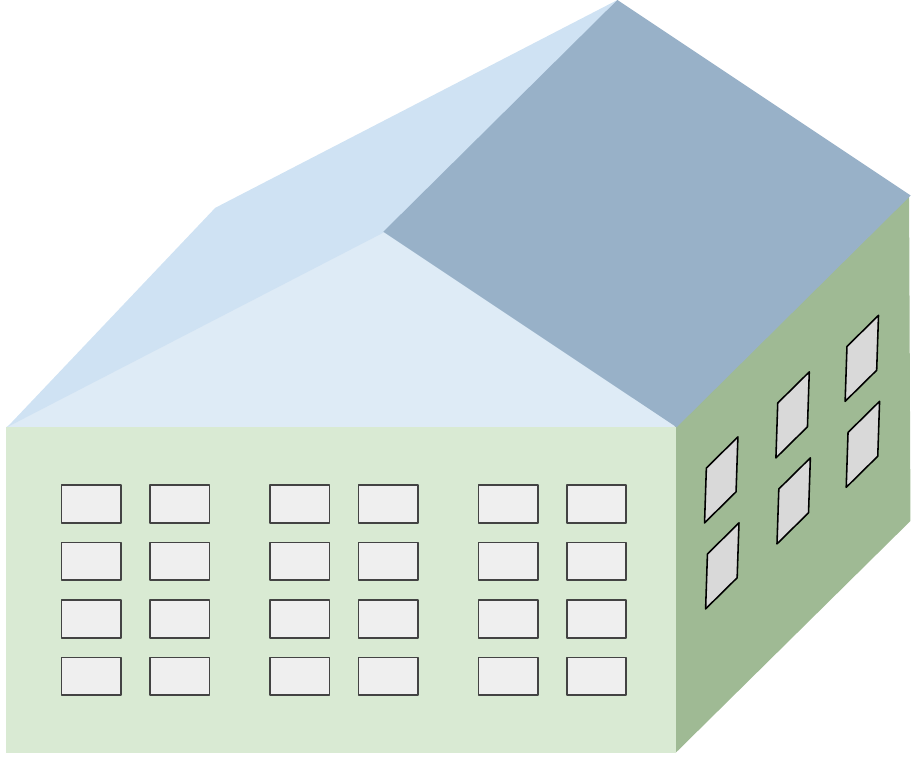}
      \caption{Gabled.}
      \label{fig:gabled_building}
    \end{subfigure}
    \caption{\sysname is designed for buildings with vertical sides and either polygonal or gabled roofs and their variations (hipped, mansard, pyramidal, skillion, \etc).}
    \label{fig:buildings}
\end{figure}

In this section, we describe the design of \sysname, beginning with an overview. 

\begin{figure}[t]
  \centering
  \includegraphics[width=0.75\columnwidth]{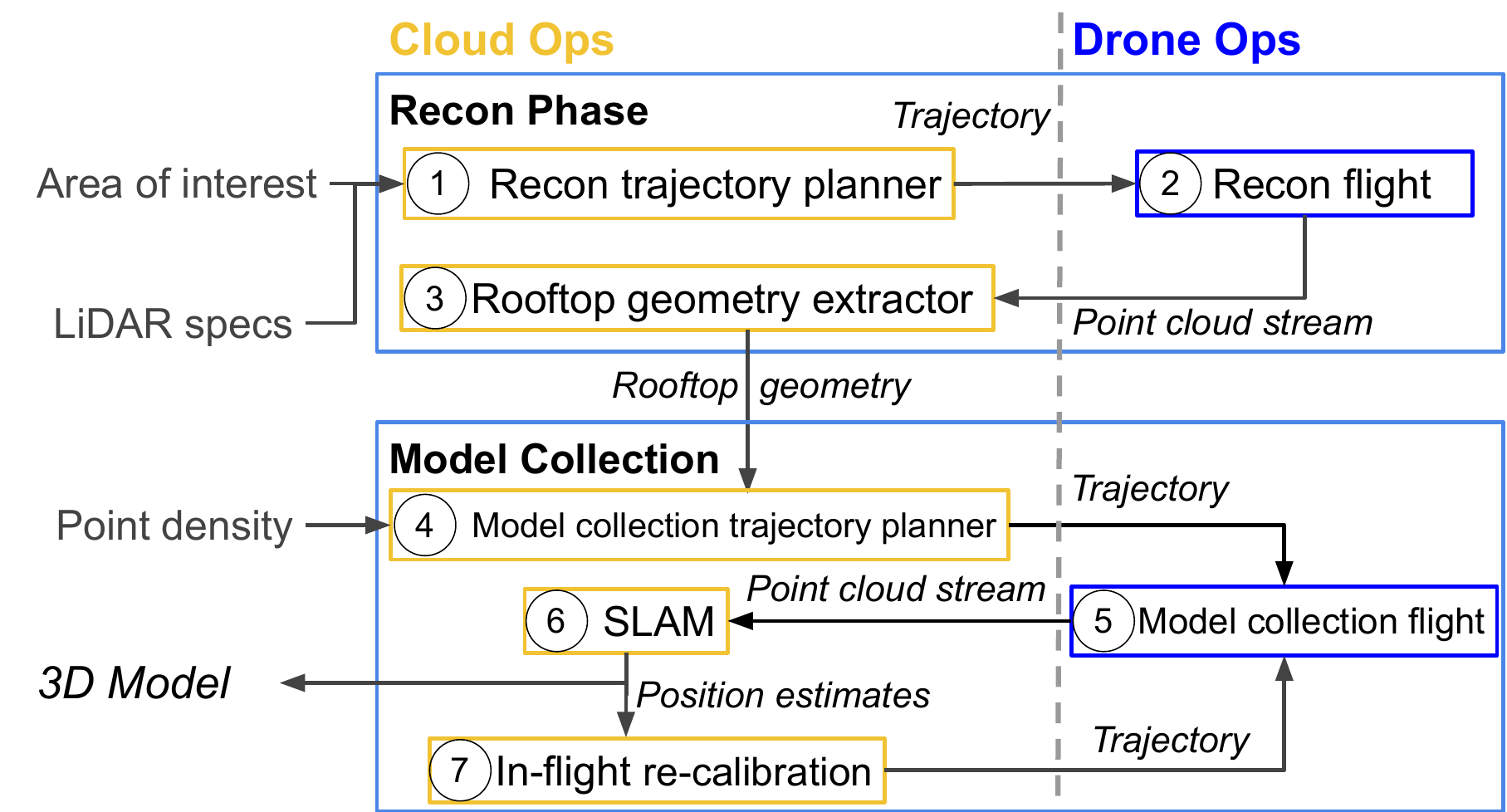}
  \caption{\mradd{\system operational model.}\label{fig:overview}
  }
\end{figure}

\subsection{Overview}

To use \sysname (\figref{fig:overview}), the user must specify: a) an \textit{area of interest}, b) the drone-mounted LiDAR specifications, and c) a minimum \textit{target point density}. Point density, the number of points per unit area \add{on the surface of a point cloud determines the level of detail of a 3D model}. 
By setting a minimum point density, \sysname ensures that all parts of the building surface are captured with at least the same minimum level of detail.
The user can adjust this parameter to balance model fidelity and battery usage, since more detailed models require longer flights.
The LiDAR specifications include the number of beams, horizontal and vertical field-of-view (FOV), and maximum range, which are all provided on a LiDAR's datasheet.

\subsubsection{Building Types}

\sysname exploits building \textit{geometry} (shape and size) to obtain high quality reconstructions. \sysname can generate 3D models for buildings with (a) vertical sides and (b) (convex or non-convex) polygonal flat roofs or gabled roofs and their variations (\figref{fig:buildings}). Variations of the gabled roof include hipped, pyramidal, skillion, and mansard roofs~\cite{roof_types}. \sysname can also reconstruct buildings where walls and roofs have outcroppings such as chimneys, HVAC systems, or windows, \etc (our evaluations include such structures in \figref{fig:mvs}, \figref{fig:slam_vs_gps_3d_model}, and \figref{fig:3d_model}).

This covers most commercial buildings as well as many residential structures. To quantify this, we used open source building geometry datasets~\cite{deeproof, buildingroof} of four cities (Witten, Ann Arbor, Manhattan, and Farmingham), to determine the percentage of buildings that \sysname can reconstruct. \tabref{tab:roof_types} shows that \sysname can reconstruct more than 99\% of the buildings found in these four cities. We leave to future work (\secref{s:conclusions}) extensions of \sysname to reconstruct rare building shapes (\eg buildings with dome roofs or large antennas) and other large physical structures (\eg aircraft and blimps). 

\subsubsection{Components}

Given these inputs, in the area of interest, \sysname guides a drone to automatically discover buildings, and constructs a 3D model of the buildings on-the-fly while \textit{minimizing flight duration} at that given minimum point density. To a first approximation, drone battery usage increases with flight duration; we have left it to future work to incorporate drone battery models (\secref{s:conclusions}). 

\sysname splits its functionality across two components: (a) a lightweight subsystem that runs on the drone (Drone Ops in~\figref{fig:overview}), and (b) a cloud-based component (Cloud Ops in~\figref{fig:overview}) that discovers buildings, generates drone trajectories, and reconstructs the 3D models on-the-fly.

To automatically discover buildings, \sysname's cloud component (\figref{fig:overview}) generates an efficient \textit{reconnaissance (recon) trajectory} \mradd{[\figref{fig:overview} (1)]} over the area of interest to discover the \textit{rooftop geometry} of buildings. The drone follows this trajectory \mradd{[\figref{fig:overview} (2)]}, \textit{streams compressed point clouds} to the cloud, extracting the geometry of the roof \mradd{[\figref{fig:overview} (3) and \secref{sec:recon}]}.

Using the rooftop geometry and the user-defined \textit{point density}, the cloud service prepares a more careful \textit{model collection} trajectory \mradd{[\figref{fig:overview} (4) and \secref{s:slam-phase}]} that designs a minimal duration flight to ensure high 3D model accuracy at the given point density. 
As the drone flies along this trajectory \mradd{[\figref{fig:overview} (5)]}, \sysname \textit{streams compressed point clouds} to the cloud component, which continuously runs SLAM \mradd{[\figref{fig:overview} (6)]} and estimates whether SLAM drift is sufficient to warrant \textit{recalibration} \mradd{[\figref{fig:overview} (7) and \secref{s:drift-est}]}. Soon after the drone completes the trajectory, the 3D model is available at the cloud service.

Below, we first describe model collection (\secref{s:slam-phase}), since that is the most challenging of \sysname's components. We then describe how \sysname extracts the rooftop geometry (\secref{sec:recon}), and conclude by describing point-cloud compression (\secref{s:compression}).

\begin{table}[t]
\small
  \caption{\sysname comparison with Photogrammetry and SLAM.}
{
  \vspace{0.1cm}
\begin{tabular}{c  c  c  c}
\hline \hline
\textbf{Property} & \textbf{Photogrammetry} & \textbf{SLAM} & \textbf{\sysname} \\ \hline 
\begin{tabular}[c]{@{}c@{}}Complexity\end{tabular} & High           & Low        & Low    \\ 
Reconstruction                                                      & Offline        & Offline    & Online \\ 
\begin{tabular}[c]{@{}c@{}}Automated trajectory planning\end{tabular} & $\times$ & $\times$ & \checkmark \\ 
\begin{tabular}[c]{@{}c@{}}Quality feedback\end{tabular}  & $\times$           & $\times$       & \checkmark    \\ 
Density control                                                     & $\times$           & $\times$       & \checkmark    \\ \hline
\vspace{-0.0cm} &
\end{tabular}
}

  \label{tab:system_comparison}
\end{table}

\subsection{Model Collection}
\label{s:slam-phase}




To build a high quality 3D model, \sysname must simultaneously (a) ensure high accuracy and completeness, (b) satisfy the minimum target point density, and (c) use short flights.
To achieve these, \sysname uses an optimization formulation that generates a minimum length trajectory that covers the building sides and roof whilst satisfying two constraints imposed by: a) SLAM, and b) target point density. 

\subsubsection{SLAM-imposed Constraints}
\label{s:slam-imposed}  

\begin{figure}
\begin{minipage}[b]{0.35\columnwidth}
\centering
\includegraphics[width=\linewidth]{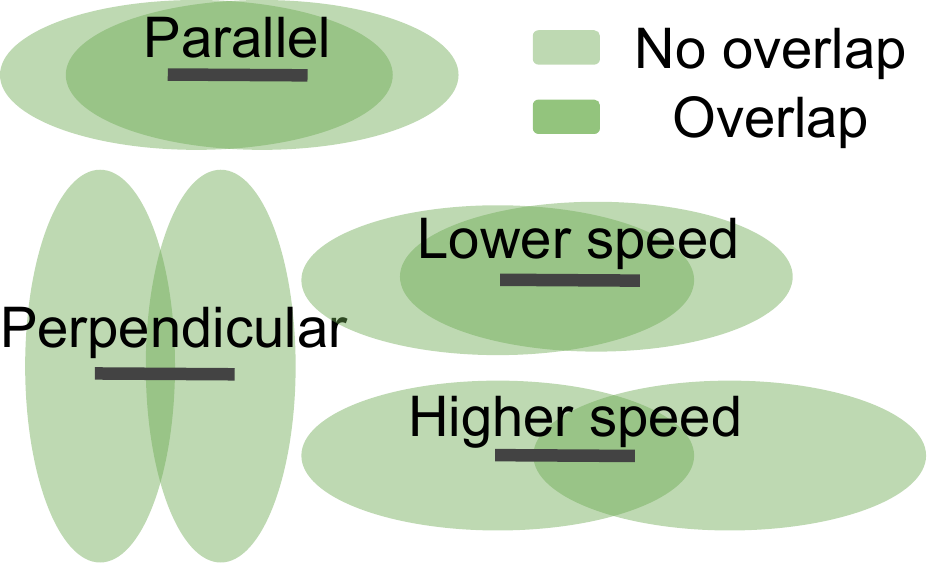}
  \caption{Overlap at various orientation and speeds.\label{fig:lidar_overlap}}
  \label{fig:overlap_example}
\end{minipage}
\hspace{0.5cm}
\begin{minipage}[b]{0.35\columnwidth}
    \centering
\includegraphics[width=\linewidth]{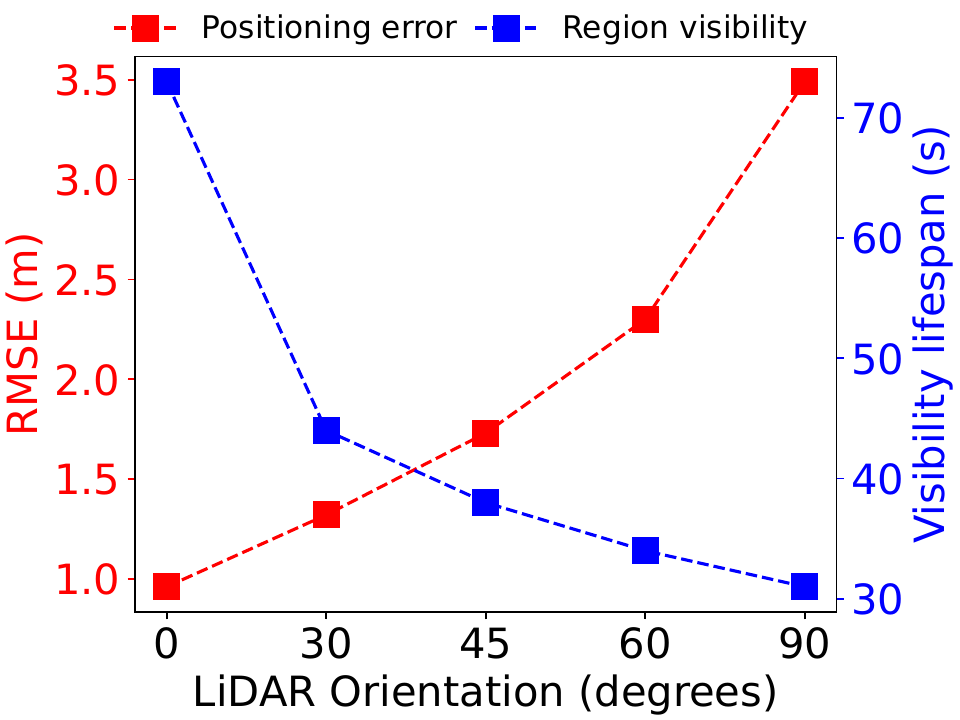}
  \caption{LiDAR's orientation Vs. SLAM positioning error.}
  \label{fig:overlap_error}
\end{minipage}
\end{figure}

The trajectory of the drone flight impacts 3D model completeness and accuracy because a poorly designed trajectory can increase SLAM error. For example, if a drone flies too fast, two successive point clouds may have little overlap (\figref{fig:lidar_overlap}), increasing SLAM error. Besides \textit{speed}, other parameters that affect SLAM error include \textit{distance from the building surface} and the \textit{orientation} of the LiDAR with respect to the ground.

SLAM algorithms are complex, so it is difficult to derive analytical models that can predict the error resulting from different choices of these parameters. Instead, we resort to a parameter sweep.
For this parameter sweep, we run SLAM in the AirSim simulator~\cite{airsim} and on real-world flight traces for different flight parameters, and find the best ones. 
We present the results of the sweep in \secref{sec:eval}, but summarize the main findings below. 
These findings point out two important factors that determine SLAM error: \textit{point density}, and \textit{degree of overlap} between successive point clouds\footnote{\mradd{SLAM estimates pose by matching successive point clouds (\secref{sec:intro}).}}.

\begin{figure}[t]
\includegraphics[width=0.9\linewidth]{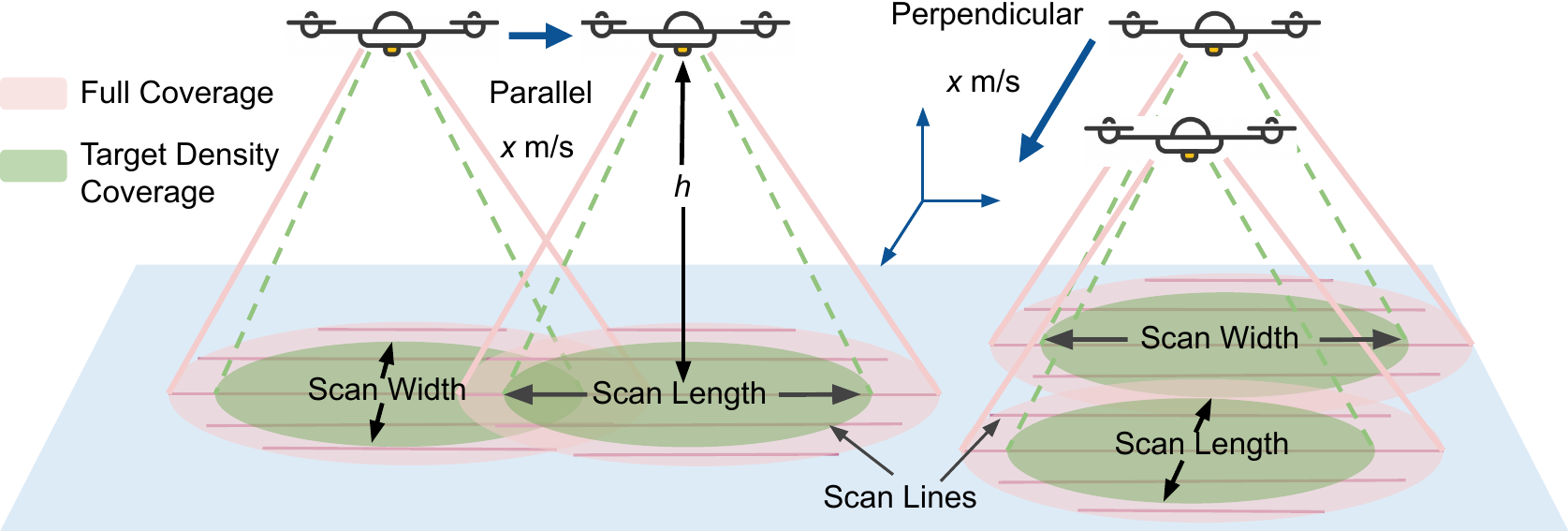}
\caption{Parallel and perpendicular LiDAR orientation.\label{fig:lidar_orientation}}
\end{figure}



Orientation impacts accuracy. 
As demonstrated in \figref{fig:slam_challenge}, on a drone, a LiDAR receives maximum returns when its scan plane is perpendicular to the ground surface.
In this position, there are multiple ways to orient the LiDAR with respect to the motion of the drone.
For instance, in a \textit{parallel} orientation (\figref{fig:lidar_orientation}), the scan lines of the LiDAR are parallel to the drone's direction of  motion.
On the other hand, the scan lines are perpendicular to the drone's motion in a \textit{perpendicular} orientation.
As shown in \figref{fig:lidar_orientation}, at a fixed height and speed, a \textit{parallel} orientation results in a higher overlap between successive point clouds as compared to a \textit{perpendicular} orientation (\figref{fig:overlap_example}).
A higher overlap results in better point cloud alignment, lower positioning error and hence higher accuracy.
\figref{fig:overlap_error}, obtained using the methodology described in \secref{sec:eval}, quantifies this intuition: different orientations have different degrees of overlap, and as overlap decreases, SLAM's positioning error increases. A parallel orientation (0$\degree$) has the lowest SLAM error because it has the highest visibility lifespan. (\textit{Visibility lifespan}, the time for which a point on the building's surface is visible during flight, is a proxy for overlap; a longer lifespan indicates greater overlap).


Speed impacts model accuracy. If the drone flies fast, two successive point clouds will have fewer overlapping points, resulting in errors in the SLAM's pose transformations and (therefore) pose estimates, which leads to poor 3D model accuracy. For high accuracy, \sysname must fly the drone slowly (\ie at 1~m/s as we demonstrate in \secref{s:data-collection}).

\begin{figure}
\begin{minipage}[b]{0.35\columnwidth}
\centering
\includegraphics[width=\linewidth]{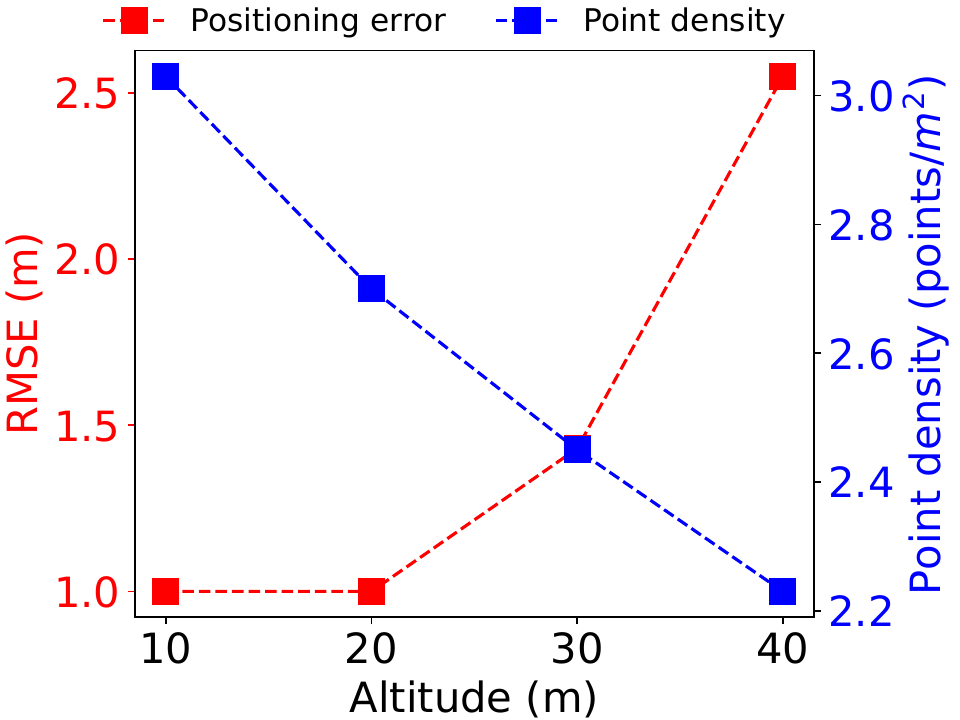}
  \caption{Point density Vs. SLAM positioning error.}
  \label{fig:density_error}
\end{minipage}
\hspace{0.5cm}
\begin{minipage}[b]{0.35\columnwidth}
\centering
      \includegraphics[width=\textwidth]{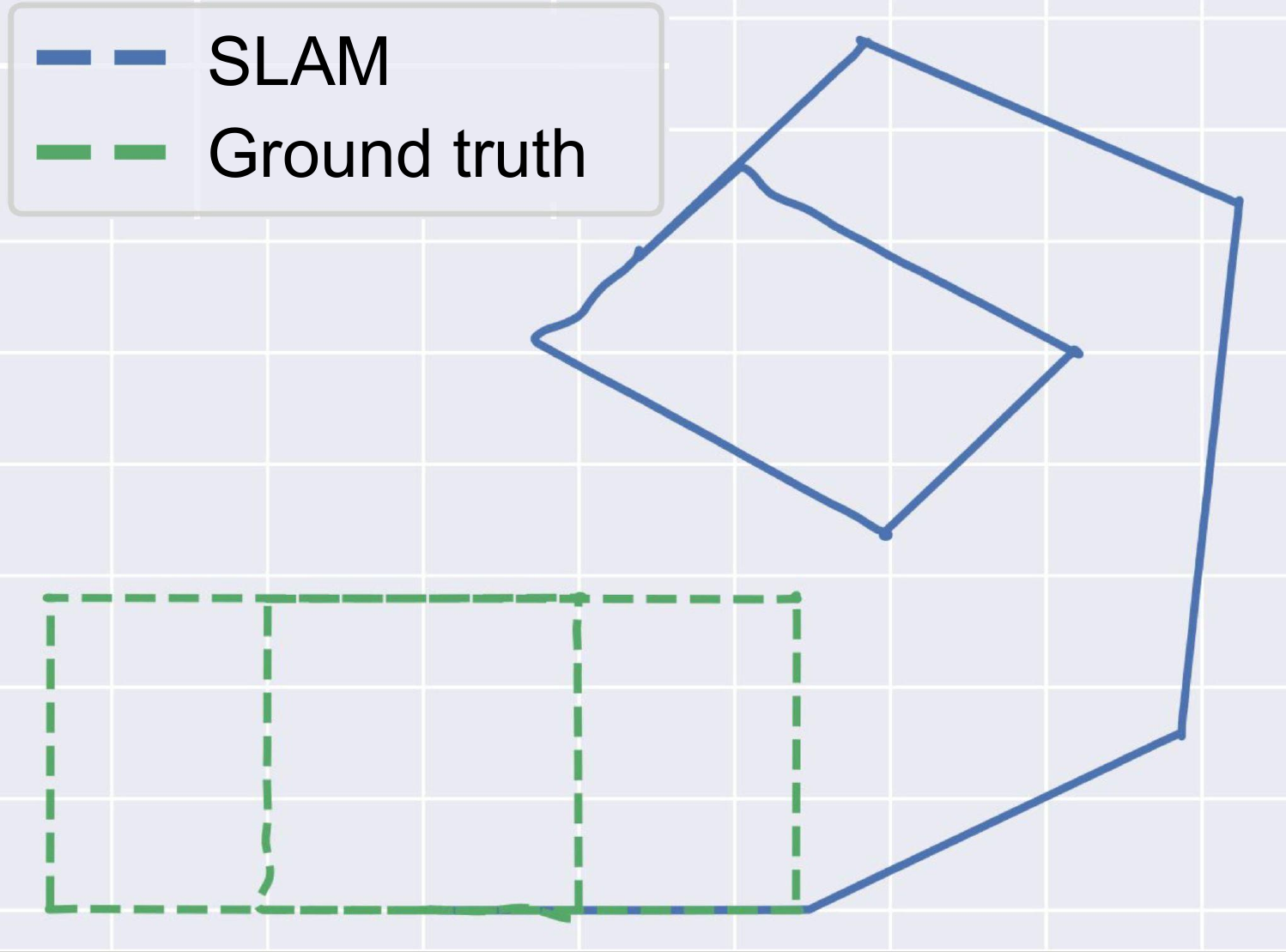}
      \caption{Rotation throws SLAM off.}
      \label{fig:slam_rotation}
\end{minipage}
\end{figure}

Height impacts both accuracy and completeness. Because LiDAR beams are radial, the higher a drone flies, the less dense the points on the surface of the building. Lower density results in worse completeness. Accuracy is also worse, because the likelihood of matching the same point on the surface between two scans decreases with point density. \figref{fig:density_error} obtained using methodology described in \secref{sec:eval} illustrates this. It plots positioning error as a function of point density by altering the altitude of the drone. The positioning errors for point densities of 2.2 points per m$^{2}$ and 3.0 points per m$^{2}$ are 2.5~m and 1.0~m respectively. So, for high accuracy, \sysname must fly the drone as low as possible (\ie no higher than 20~m above the surface, as we demonstrate in \secref{s:data-collection}).





Finally, the drone must \textit{\textbf{never rotate}} the LiDAR. This maneuver can increase SLAM error. \figref{fig:slam_rotation} shows an example in which the green dashed line is the ground truth drone trajectory, and the blue line SLAM's estimated pose. In the bottom right corner, when the drone rotates, SLAM is completely thrown off.

Two of these constraints are universal: that the parallel orientation is best, and that the drone must never rotate the LiDAR. Two others (height and speed) depend on the LiDAR characteristics and the SLAM algorithm. For a different model of LiDAR than the one we have used in this paper (a 64-beam Ouster) or SLAM implementation (Cartographer), we may need to re-run the simulations to obtain these flight parameters. We envision a practical implementation of \sysname will include pre-computed parameters for different LiDAR models, so users will not have to determine these.

\subsubsection{Point Density-imposed Constraints}
\label{s:density-imposed}

As the drone flies, its LiDAR obtains points on the surface of the building at successive instants. Let the term \textit{scan} denote the portion of the surface for which the LiDAR's points have a density no less than the target point density (the green area in \figref{fig:lidar_orientation}). For a given orientation, scan length is measured along and scan width perpendicular to the drone's motion (\figref{fig:lidar_orientation}). Given a distance $h$ at which the drone flies from the surface, the target point density requirement imposes constraints on \textit{scan width}. Two successive scans of the surface (\eg legs of a U-shaped trajectory) cannot be separated by more than the \textit{scan width} without violating the density constraint.

\sysname's key observation is that these constraints can be derived from an analytical model of the LiDAR, given its configuration. For instance, for an Ouster LiDAR with 64 beams, a vertical field of view of 45$\degree$, two consecutive beams are separated by 0.7$\degree$ ($\frac{45}{64}$). During one full 360$\degree$ rotation, the laser emits 1024 \textit{pulses}; successive pulses are 0.35$\degree$ ($\frac{360}{1024}$) apart. Each pulse generates a point in the point cloud\footnote{This is idealized. In practice, LiDARs may drop some reflections if they are noisy~\cite{Lidarsim}. So, our density guarantee is \textit{nominal}. Future work can model this noise for better equi-dense trajectory designs.}. Then, the 3D coordinates of a point from the $n$-th pulse of the $b$-th beam from a LiDAR mounted on a drone at height $h$ pointing downwards is:
\begin{equation}
\left( x, y, z \right) =
\
\left( 
r \sin \theta_{L},
r \cos \theta_{L} \sin \theta_{B}, 
r \cos \theta_{L} \cos \theta_{B}  
\right)
\end{equation}
where ($ \theta_{L} = \frac {\theta_{F}}{n} $) is the angle of the $n$ pulse of 
the $b$-th beam whose 
beam angle is $\theta_{B} = \frac{b}{360}$, $r_{max}$ is the maximum range,
$\theta_{F}$ is the vertical field of view,
and $r$ is the distance between the LiDAR and where the pulse hits the surface:
\begin{equation}
\label{equation:lidar}
r = \frac{h}{\cos \left( \theta_{L} + \theta_{B} \right) }, \quad \quad \forall r <= r_{max}
\end{equation}
From this, \sysname derives the coordinates of every point on the surface. From this point cloud, it can derive the fine-grained point density distribution. Thus, given a minimum point density distribution and a distance $h$ (obtained from the SLAM-derived constraints), \sysname calculates the scan width. Using the same procedure, \sysname derives the \textit{scan length}, the length on the surface along the direction of the flight containing points from a single frame satisfying the minimum density constraint (\figref{fig:lidar_orientation}). Scan length helps discretize the search space for trajectory optimization.

\begin{figure*}[t]
\captionsetup[subfigure]{labelfont=rm}
    \centering
    \begin{subfigure}{0.16\linewidth}
      \includegraphics{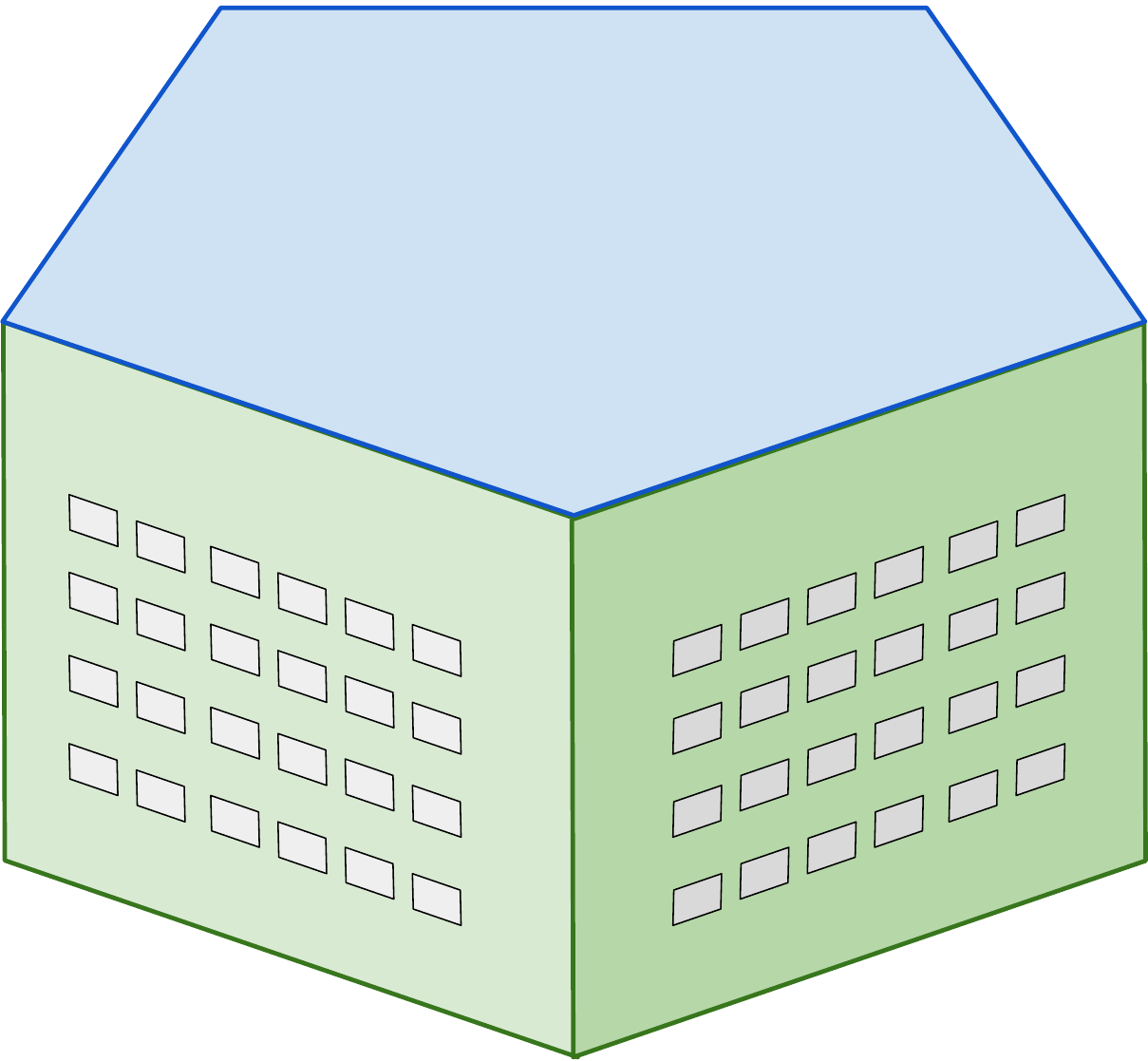}
      \caption{Pentagon building.} 
      \label{fig:pentagon_tall}
    \end{subfigure}
    \begin{subfigure}{0.16\linewidth}
      \includegraphics{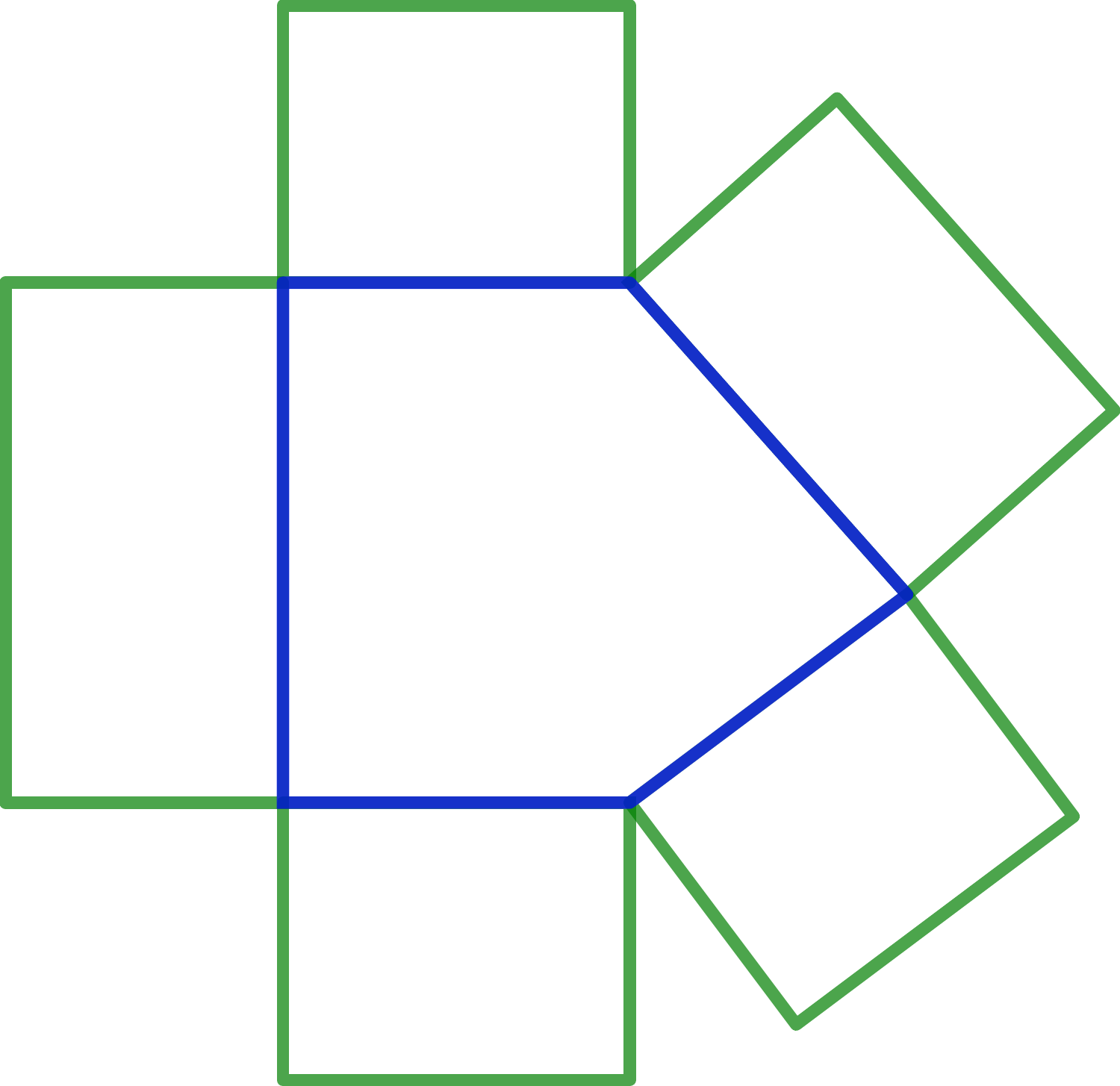}
      \caption{Unfolded pentagon.}
      \label{fig:pentagon_unfolded}
    \end{subfigure}
    \begin{subfigure}{0.16\linewidth}
      \includegraphics{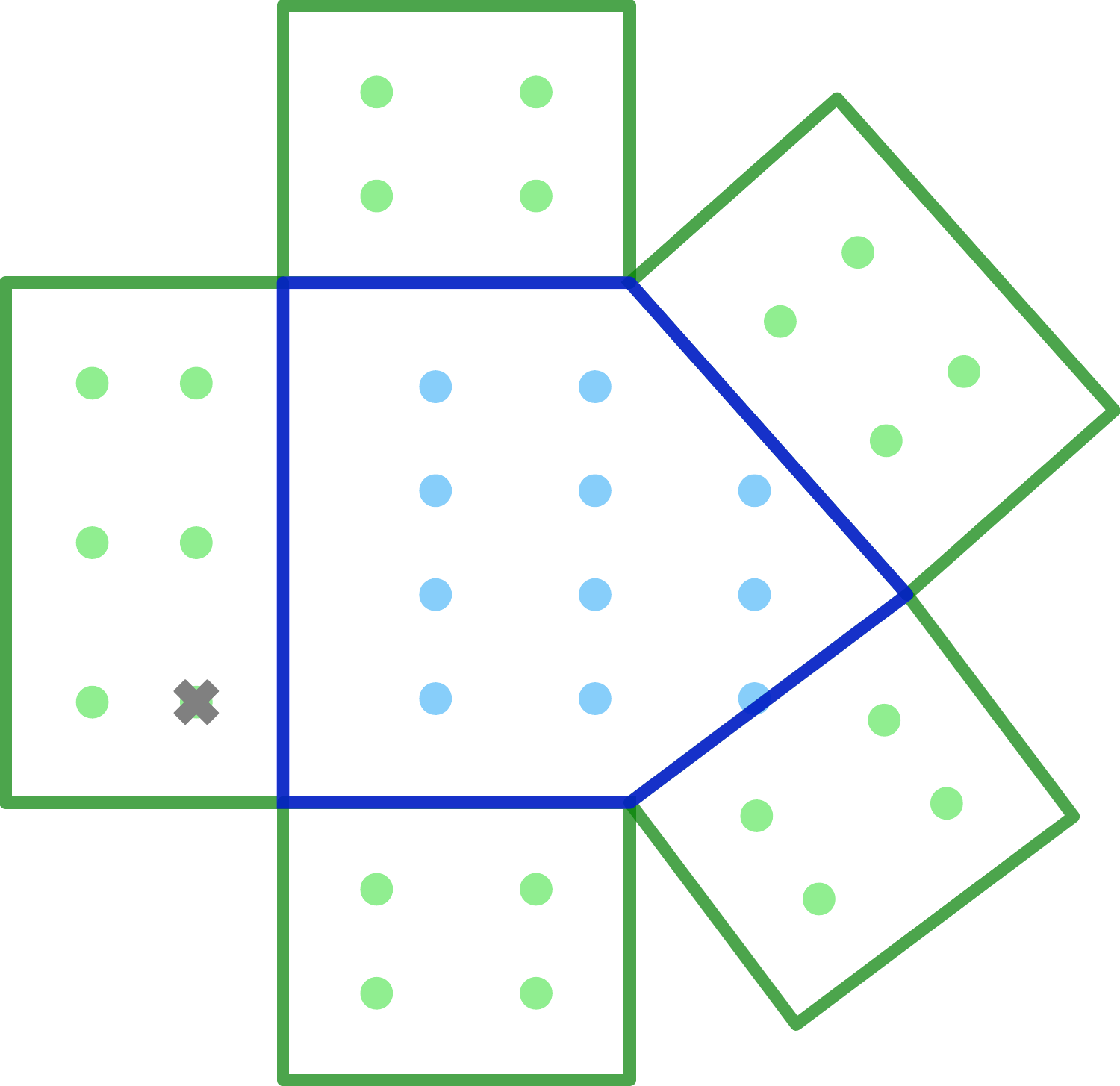}
      \caption{Meshed pentagon.}
      \label{fig:pentagon_mesh}
    \end{subfigure}
    \begin{subfigure}{0.16\linewidth}
      \includegraphics{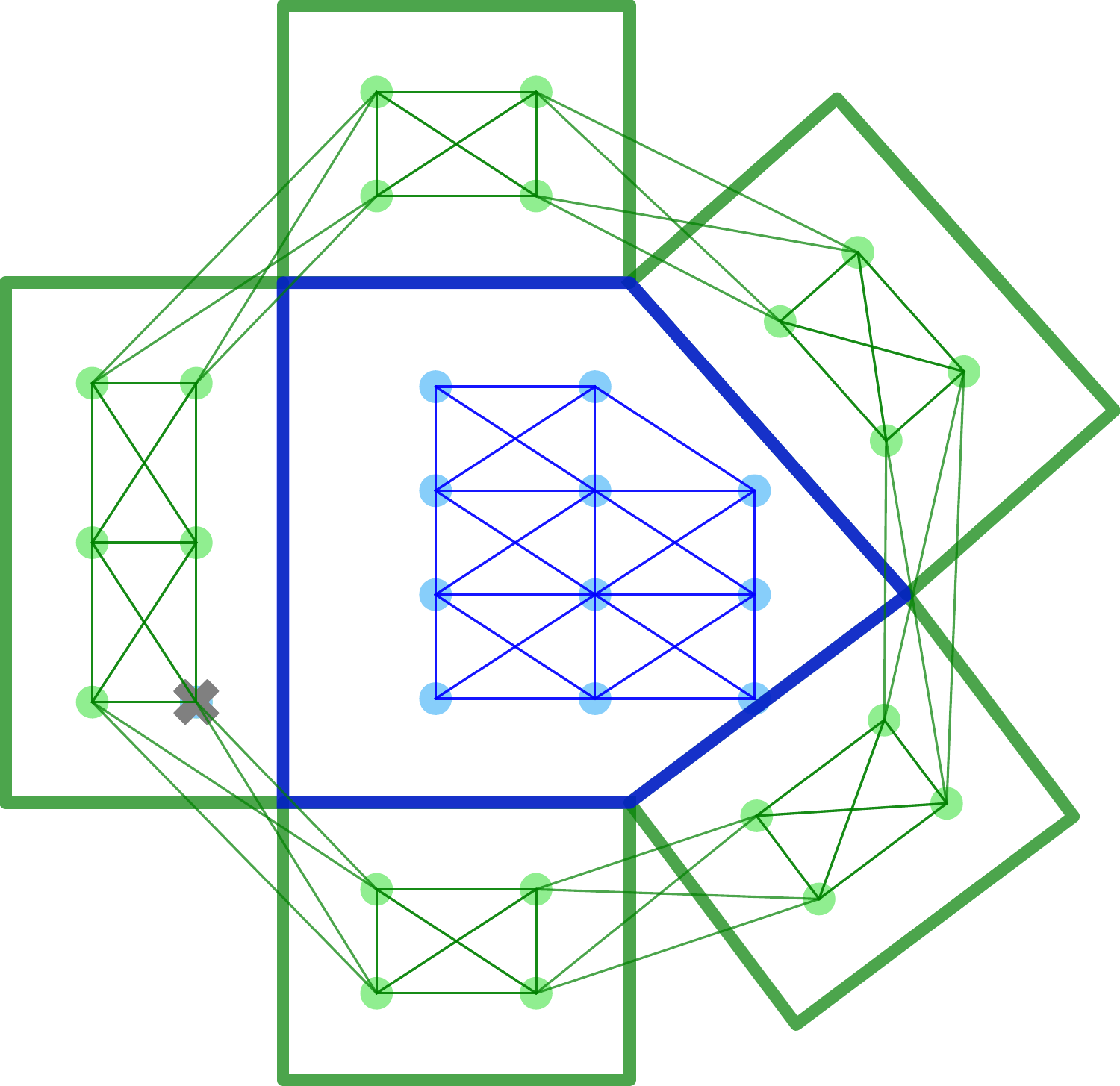}
      \caption{Graph of pentagon.}
      \label{fig:pentagon_graph}
    \end{subfigure}
    \begin{subfigure}{0.16\linewidth}
        \includegraphics{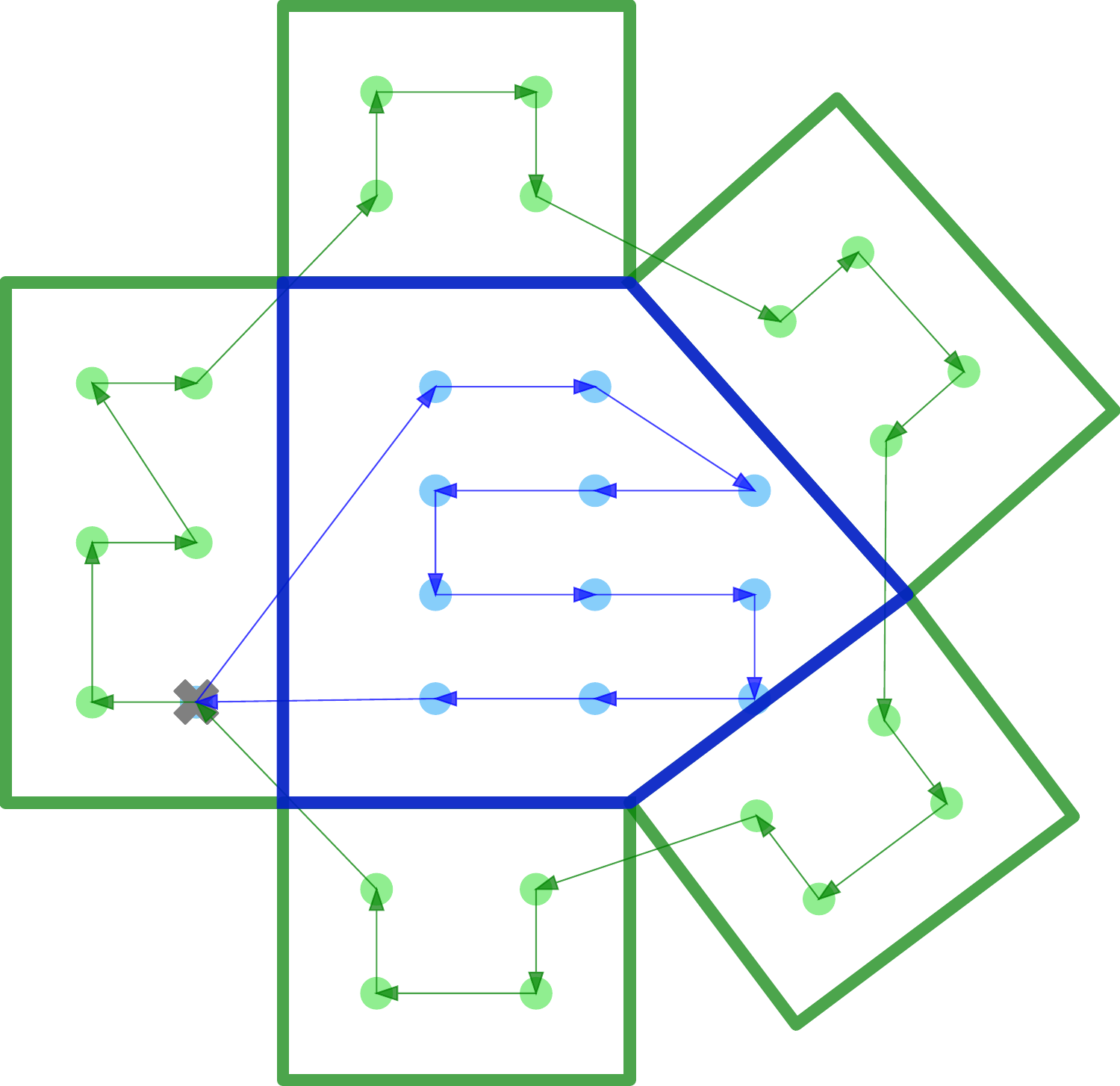}
        \caption{Derived trajectory.}
        \label{fig:pentagon_path}
    \end{subfigure}
    \begin{subfigure}{0.16\linewidth}
        \includegraphics{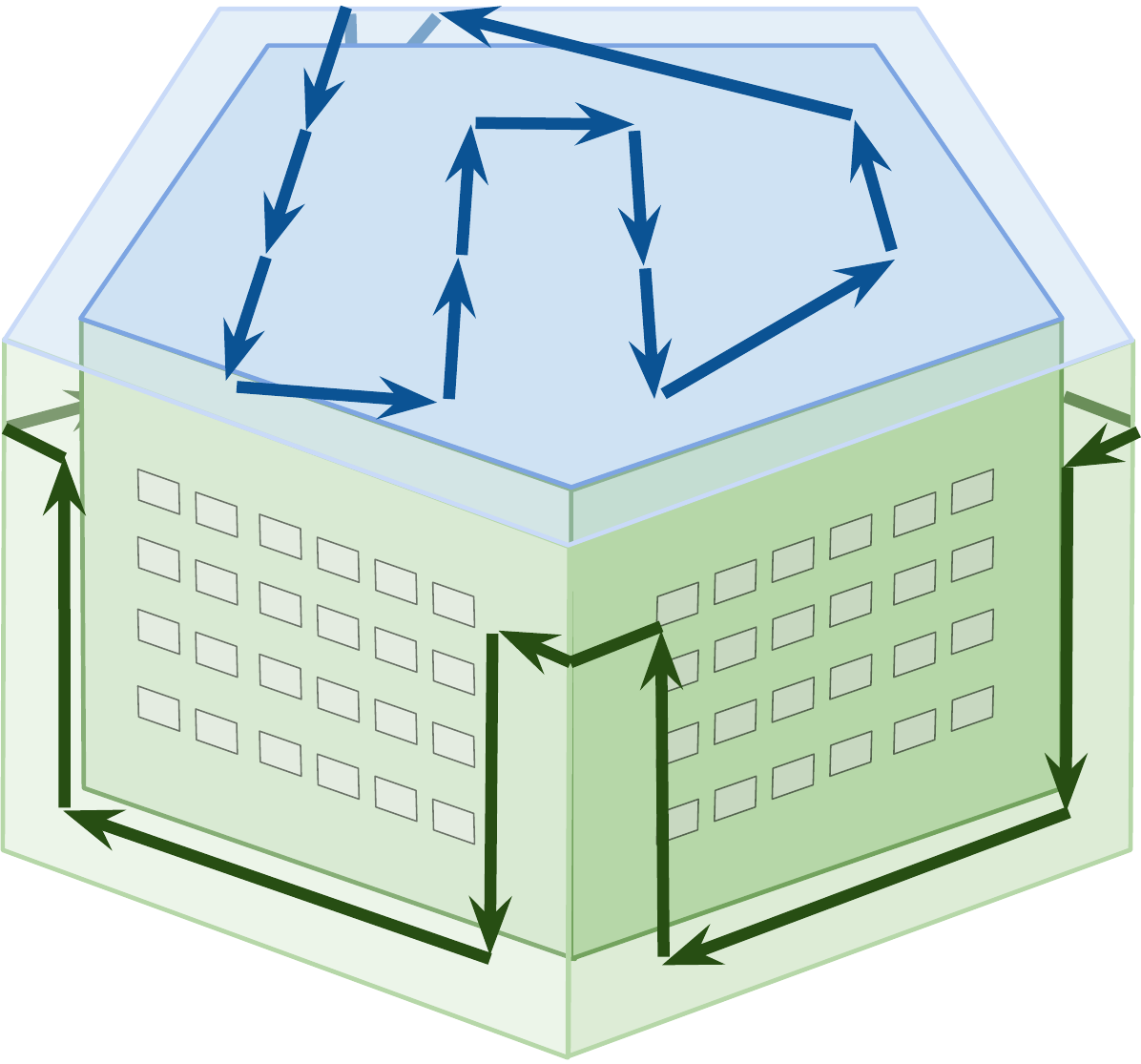}
        \caption{Folded trajectory.}
        \label{fig:pentagon_folded}
    \end{subfigure}
    \caption{Illustrating steps in optimized trajectory generation.}
    \label{fig:pentagon_trajectory}
\end{figure*}

\subsubsection{Optimized Trajectory Generation}
\label{s:opt-traj}
\sysname uses a novel optimization formulation to generate the shortest flight path around the building.
Inspired by prior work~\cite{cpp-grid1,ilp-karthik}, it uses an Integer Linear Programming (ILP) based optimization to generate the shortest flight path that respects point density and SLAM-imposed constraints for the class of buildings discussed above. \figref{fig:pentagon_trajectory} illustrates the steps, described below, in trajectory generation for a building with a flat pentagonal roof (\figref{fig:pentagon_tall}).


\textbf{Unfolding.} Unfolding a 3D building on a 2D plane reduces trajectory planning to coverage path planning (CPP)~\cite{cpp}. \sysname unfolds the sides of the building, so that the resulting object is planar. In \figref{fig:pentagon_unfolded}, this results in a pentagon surrounded by rectangles. (\sysname applies a similar unfolding trick to the gabled roof and its variations, details omitted for brevity.)

\textbf{Meshing.} This imposes a rectangular grid on the roof and all sides to discretize the unfolded structure into grid elements  with dimensions \textit{scan length}/2 by \textit{scan width}/2 (the points in \figref{fig:pentagon_mesh} depict the centers of the rectangles). Discretization reduces the search space~\cite{cpp-grid1, cpp-grid2}, and using the scan width ensures target point density. The longer side of the grid element parallels the longer side or dominant orientation of the corresponding polygon; \sysname aligns the LiDAR parallel to the longer side to maximize flight segments with a parallel orientation. Finally, a traversal that visits all mesh centers guarantees coverage of the structure. For a non-convex flat roof, \sysname generates the mesh on the convex hull of the roof.


\textbf{Graph Generation.} \sysname now embeds a graph on the mesh; the vertices are the centers of rectangles, and each center is connected to every neighbor with an edge (\figref{fig:pentagon_graph}). \sysname actually embeds two sub-graphs: one on the roof mesh, and one on the side meshes taken together. It generates \textit{tours} separately for each sub-graph, because these two require different LiDAR orientations. For each sub-graph, the tours start and terminate at a fixed origin (denoted by $\times$ in \figref{fig:pentagon_mesh}), which ensures a SLAM loop closure before the LiDAR needs to be rotated for the other sub-graph. 


\textbf{Optimization Formulation.} For each sub-graph, \sysname models the trajectory as a Travelling Salesman Problem (TSP) tour, starting and ending at the origin. It formulates the TSP traversal as an ILP. This step takes a subgraph $G = (V, E)$ as input, where $V$ and $E$ are the set of vertices and edges, respectively and vertices indexed $i = 1, 2, ..., N$, with $i=1$ the origin. Each edge $(i, j) \in E$ has a weight $w_{ij}$ which is the 2D Euclidean distance between the position of the vertices $i$ and $j$. If the drone flies from vertex $i$ to vertex $j$, given $(i, j) \in E$, the binary decision variable $x_{ij}$ set to 1, 0 otherwise. The formulation below finds a minimum length trajectory:
\begin{mini!}
    {}{\sum_{i} \sum_{j} \big(w_{ij}x_{ij} + \lambda p_{ij}\big)}
    {\label{eq:opt}}{}
    \addConstraint{\sum_{\forall i} x_{ij}}{= 1, \quad}{\forall j \neq 1}
    {\label{eq:const1}}{}
    \addConstraint{\sum_{\forall i} x_{ij}}{\geq 1, \quad}{j = 1}
    {\label{eq:const2}}{}
    \addConstraint{\sum_{\forall i \neq j} x_{ij}}{= \sum_{\forall k \neq j} x_{jk}, \quad}{\forall j}
    {\label{eq:const3}}{}
    \addConstraint{u_1 = 1 \quad}{\text{and} \quad 2 \leq u_i \leq N, \quad}{\forall i \neq 1}
    {\label{eq:const4}}{}
    \addConstraint{u_i - u_j + 1}{\leq (N-1)(1-x_{ij}), \quad}{2 \leq i \neq j \leq N}
    {\label{eq:const5}}{}
\end{mini!}
The first term in the objective is the tour length. To force tours to traverse the structure in a parallel orientation as much as possible, the second term penalizes edges selected in non-parallel direction by a factor $\lambda \geq 0$, where $\lambda$ is a hyperparameter. If a selected edge $x_{ij} = 1$ is in non-parallel direction $p_{ij}$ is set to 1, 0 otherwise.

The optimization is subject to several constraints: (\textbf{\ref{eq:const1}}, \textbf{\ref{eq:const2}}) the drone visits each vertex of the graph exactly once, except the origin; (\textbf{\ref{eq:const3}}) if a drone visits a vertex $j$, it has to leave the same vertex; (\textbf{\ref{eq:const4}}, \textbf{\ref{eq:const5}}) no subtours of size $< N$ are allowed. We use Miller-Tucker-Zemlin based subtour elimination constraint~\cite{tsp-mtz} for this optimization. It requires auxiliary decision variables $u_i$ for each vertex.


\textbf{Refolding.} Once it obtains the 2D trajectories (Fig.~\ref{fig:pentagon_path}), the trajectory planner projects each 2D point to 3D coordinates. Since the drone has to fly at a distance $h$ from the building, \sysname obtains the 3D coordinates at a distance of $h$ from the scanned face by projecting each point on the plane parallel to the face at a distance $h$ away from it (Fig.~\ref{fig:pentagon_folded}).

\subsection{Drift Estimation and Re-calibration}
\label{s:drift-est}

Although \sysname generates a careful model collection trajectory designed to minimize positioning error, SLAM accumulates error on long flights. 
To minimize drift error, SLAM uses \textit{loop closure}~---~this return to a previously-visited spot allows SLAM to re-calibrate itself. 
However, it is hard to predict \textit{when} SLAM incurs significant drift, so \sysname continuously estimates drift and triggers a mid-flight return to \textit{origin}, a designated spot at which to close the loop, and restart a new SLAM session. 
Mid-flight re-calibration is the primary reason for \sysname's cloud offload of SLAM, because, to detect drift while the drone is flying, it must obtain SLAM's position estimates to assess if there is a drift. 
Detecting excessive drift is a non-trivial problem since \sysname has no way of knowing when SLAM's position estimates are wrong because it does not have accurate ground truth.

\begin{algorithm}
    \caption{Detecting excessive drift.}
    \label{algo:imperfection_detection}
    \begin{algorithmic}[1]
    
    \Function{DetectExcessiveDrift}{$S, G$} \Comment{$S$: SLAM poses, $G$: GPS Tags}

    \State $ \hat{S}, \hat{G} = \Call{TimeSynchronization}{ {S}, {G}}$
    \State $ \hat{G_{a}} = \Call{GPSToMercator} { \hat{G} } $ 
    \State $ t_{cw} = \Call{UmeyamaAlignment} { \hat{G_{a}}, \hat{S} } $ 
    \State $ \hat{S_{a}} = \Call{TransformTrajectory} { \hat{S}, t_{cw} } $ 
    \For { $ \hat{{s}_{a-i}} : \hat{S_{a}} ; \hat{g_{a-i}} : \hat{G_{a}} $ }
        \State $r_{i} = \Call{RMSE} { \hat{s_{a-i}}, \hat{g_{a-i} } } $
        \If { $ \Call {IsExcessive} { r_{i} } $ } 
            \State $I.\Call{Append}{g_{a-i}}$
        \EndIf
    \EndFor
    \Return $I$ \Comment{$I$: Imperfect regions}
    \EndFunction
    \end{algorithmic}
    \end{algorithm}

\sysname's key insight is to detect drift by comparing the \textit{shape} of SLAM's estimated trajectory (see \figref{fig:slam_rotation} for an example) with the \textit{shape} of the GPS trajectory. GPS does not suffer from drift error, and this approach is robust to GPS errors, since it matches larger segments of the two trajectories, not their precise positions. 
Specifically, \system continuously executes 3D SLAM on the stream of compressed LiDAR frames from the drone, and estimates the pose of each frame. 
Then, it runs the algorithm described in \algoref{algo:imperfection_detection}. 
It transforms GPS readings using the Mercator projection (line 2). 
Then, it aligns the GPS trajectory and the SLAM-generated trajectory using the Umeyama algorithm~\cite{umeyama} (line 3) to determine the rigid transformation matrices (\ie translation, and rotation) that best align SLAM and GPS poses (line 4). \system partitions trajectories into fixed length segments and after alignment, computes the RMSE between the two trajectories in each segment, and uses these as an indicator of excessive drift (lines 5-11): if the RMSE is greater than a threshold $\rho$, \system invokes return-to-origin (\figref{fig:recalibration}).

\subsection{Discovering Rooftop Geometry}
\label{sec:recon}
\sysname's model collection requires the rooftop geometry as input. Depending on the type of structure, it may be difficult to get this information. To estimate this information, \sysname uses a reconnaissance (recon) flight over the area of interest.
This recon flight must be as short as possible, to minimize battery usage.
In addition, recon (a) must not assume prior existence of a 3D model of the building (prior work in the area makes this assumption~\cite{lidarbuildingdetection, lidarimagebuildingdetection}); (b)  must be robust to nearby objects like trees that can introduce error; and (c)  must generalize to the class of buildings \sysname targets (\secref{sec:design}).

\subsubsection{The Recon Trajectory} 
\begin{figure*}[t]
  \begin{minipage}{0.5\columnwidth}
    \centering
 \includegraphics[width=0.7\textwidth]{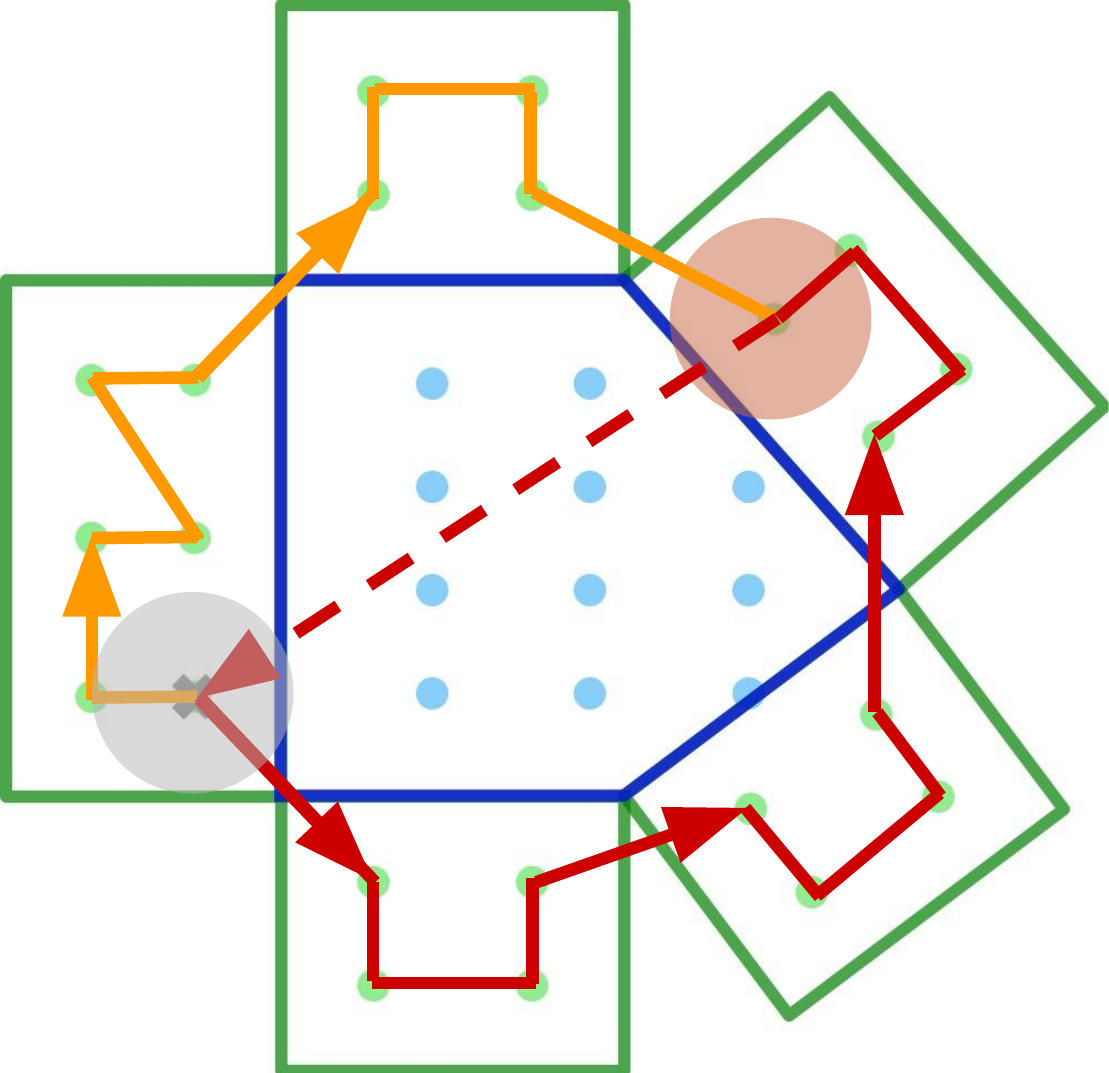}
      \caption{\add{\sysname re-calibration flight of a pentagon-shaped building (central blue region is the roof, and green regions are the unfolded sides). The initial model collection flight for the sides of the building (red line) detects drift (red circle), and then initiates a return-to-origin maneuver (dotted red line). Then, a re-calibration flight (orange line) from the origin (grey circle) collects 3D point clouds from the remaining unscanned regions.} 
      } 
      \label{fig:recalibration}
  \end{minipage}
\hfill
\begin{minipage}{0.48\columnwidth}
  \centering
  \includegraphics[width=0.6\textwidth]{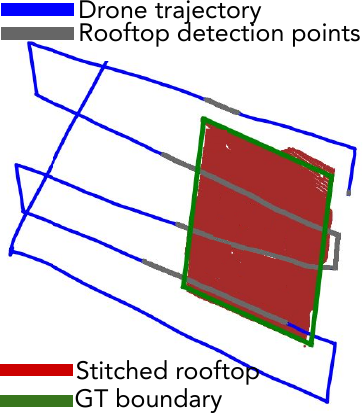}
  \caption{\sysname's building detector on a real 40m~x~70m~x~20m building.}
  \label{fig:boundary}
\end{minipage}
\end{figure*}

Recon uses a rectilinear scan (blue line in \figref{fig:boundary}), but unlike model collection, during recon the drone flies \textit{fast} (4~m/s) and \textit{high} (60~m above the building's roof\footnote{We assume the nominal building heights in an area are known, for example, from zoning restrictions.}), with the LiDAR mounted in a \textit{perpendicular} orientation in order to have the shortest duration flight possible (we justify these flight configurations in \secref{sec:eval:building}). During this flight, \sysname streams point clouds to its cloud component, which runs the \textit{boundary detection} algorithms described briefly below.

\subsubsection{Surface Extraction} The cloud component receives GPS-tagged compressed point clouds from the drone. It first uncompresses them, then computes the \textit{surface normal} of every point in the point cloud. A surface normal to a point determines the direction normal to the surface formed by points within a fixed radius of the point. Calculating surface normals is compute-intensive, so \sysname offloads this operation to the GPU.  Then, \sysname uses RANSAC~\cite{ransac} (a plane-fitting algorithm) to segment the LiDAR points into groups of points that fall onto planes. It removes the plane furthest from the drone (likely to be the ground plane) and eliminates planes with irregular surface normals (\eg from nearby trees). The remaining planes correspond to rooftops. For additional robustness, it uses majority voting across multiple frames to identify the rooftop.

\subsubsection{Estimating the Boundary}
\sysname uses the drone's GPS location to transform each surface to the same coordinate frame of reference, then combines all surfaces into a single point cloud that represents the extracted rooftop of the building (red rectangle in \figref{fig:boundary}). 
To extract the boundary of the building, it extracts the \textit{alpha shape}~\cite{alphashape} (a sequence of piece-wise linear curves in 2D) of the stitched point cloud (green outline in \figref{fig:boundary}). This allows \sysname to generalize to non-convex shapes as well. Finally, to detect the boundary of multiple buildings, \sysname clusters the rooftop point clouds.


\subsection{Point-Cloud Compression}
\label{s:compression}

LiDARs generate voluminous 3D data. For instance, the Ouster OS1-64, with 360$\degree$ horizontal and 45$\degree$ vertical field-of-view (FoV), generates 20 point clouds per second requiring 480~Mbps, well beyond the capabilities of today's cellular standards. \sysname compresses these point clouds to a few Mbps (1.2 to 4.0), using two techniques: viewpoint filtering, and octree compression.

Viewpoint filtering removes returns from beams directed towards the sky, from objects beyond the LiDAR range (because they generate \textit{zero returns}) and also from the drone's body. \system compresses the retained data using an octree compression algorithm~\cite{octree} designed for point clouds (so better than data-agnostic compression techniques like Gzip). It uses different configurations to control octree and point resolution that govern compressibility, to achieve point-cloud transmission rates of 1.2, 2.5, and 3.8~Mbps (\secref{sec:eval}) corresponding to high, medium and low compression, respectively (all within achievable LTE speeds).

\section{Evaluation}
\label{sec:eval}


We have implemented \sysname using the Point Cloud Library (PCL~\cite{octree}), the Cartographer~\cite{Cartographer} LiDAR SLAM implementation, the Boost C++ libraries~\cite{Boost}, and the Robotic Operating System (ROS~\cite{ros}). For the recon phase described in \secref{sec:recon}, we used functions from the Point Cloud Library (PCL~\cite{octree}) for plane-fitting, outlier removal and clustering. Our compression and extraction modules are ROS nodes that use PCL. The drift detection module uses a Python package for the Umeyama alignment~\cite{grupp2017evo}. The trajectory optimization uses Gurobi~\cite{gurobi}. Not counting libraries and packages it uses, \sysname is 16,500 lines of code. 

We use a photorealistic simulator, AirSim~\cite{airsim} that models realistic physical environments using a game engine, then simulates drone flights over these environments and records sensor readings taken from the perspective of the drone. AirSim has a parametrizable model for a LiDAR; we used the parameters for the Ouster OS1-64 in our simulation experiments. \sysname generates trajectories for the AirSim drone, then records the data generated by the LiDAR, and processes it to obtain the 3D model. For computing the metrics above, we obtain ground truth from AirSim. To build the ground truth 3D model, we flew a drone equipped with a LiDAR several times over the region of interest in AirSim (using exhaustive flights) and then stitched all the resulting point clouds using ground truth positioning from AirSim.

In addition, we have collected data from nearly 30 flights (each of about 25 minutes) on an M600Pro drone with an Ouster OS1-64 LiDAR on a commercial complex. For almost all experiments, we evaluated \sysname on both \textit{real-world} and simulation-driven traces. Simulation-driven traces give us the flexibility to explore the parameter space more (as we show below). However, we use real-world traces to validate all these parameter choices and estimate reconstruction accuracy in practice. For real-world experiments, \add{we offload to an AWS VM with 16 cores, 64 GB RAM and an Nvidia T4 GPU}. \mradd{The AWS VM was located approximately 3,900~km from the drone.}



We quantify end-to-end latency, 3D model accuracy and completeness (\secref{s:slam-phase}), and positioning error. We also quantify \sysname's energy-efficiency (using flight duration as a proxy for drone battery usage) and the computational capabilities of its processing pipeline. Flight duration includes recon and model collection.

\subsection{3D Model Reconstruction}
\label{sec:eval:accuracy}



\subsubsection{Comparison: Photogrammetry} 

We compare \sysname against ColMap~\cite{colmap}, a state-of-the-art photogrammetry-based tool that uses multi-view stereo (MVS~\cite{furukawa15:_multi_view_stereo}). ColMap is extensively used in the vision community as the gold standard for 3D reconstruction~\cite{martin2021nerf, liu2022depth}.
For this, we use two rectangular buildings: a) a \textit{large} 100m~x~50m~x~20m (L~x~W~x~H) and, b) a \textit{small} 50m~x~50m~x~20m building in AirSim. 
\mradd{
We compare both approaches using three metrics: a) accuracy, b) completeness, and c) end-to-end reconstruction time (proxy for latency).
}
We calculated the accuracy and completeness of the models generated by these approaches by comparing them against ground truth models generated from AirSim. Lower is better for accuracy and completeness. 
\mradd{
In addition to these, we measured the end-to-end reconstruction time, the total time taken to construct the 3D model end-to-end.
End-to-end reconstruction time is a function of flight time, and processing time.
}
Flight time reports the model collection flight duration.
Processing time indicates the time to construct a 3D model with the gathered data.


For these experiments, \sysname uses \textit{compressed point clouds} with bandwidth requirements that are compatible with LTE speeds today (\ie upload bandwidth of 3.8~Mbps); we study the effect of compression on \sysname model reconstruction more in \secref{s:ablation-study}. For ColMap, we flew the drone in various trajectories recommended for high-quality reconstruction~\cite{dronedeploy} and collected 2D images for \textit{offline} reconstruction; we report the result for the best performing trajectory. For a more than fair comparison, in ColMap, we assume the building location and roof geometry is known a priori. Without this assumption, 
ColMap's reconstruction quality is poor and its reconstruction and flight times are much longer.

\begin{table}[t]
\caption{\add{Reconstruction accuracy (acc.), completeness (compl.), flight time, processing time and end-to-end reconstruction time for two buildings using: a) photogrammetry reconstruction with an optimized trajectory (ColMap), and b) \sysname.}}
\vspace{0.1cm}

\small{
\centering
\begin{tabular}{c c c c c c}
\hline \hline
\textbf{Scheme} & \textbf{Acc. (m)} & \textbf{Compl. (m)} & \textbf{Flight (s)} & \textbf{\add{Processing (s)}} & \textbf{\add{End-to-end reconstruction (s)}} \\ \hline
\multicolumn{6}{c}{ \textit{Large building (100~m~x~50~m~x~20~m)}} \\ \hline
\begin{tabular}[c]{@{}c@{}}ColMap\end{tabular}    & 0.16          & 0.75          & \textbf{1320} & \add{31600} & \add{32900} \\ 
\begin{tabular}[c]{@{}c@{}}\sysname \end{tabular}        & \textbf{0.09} & \textbf{0.05} & 1430 & \textit{\add{in-flight}} & \textbf{\add{1430}} \\ \hline

\multicolumn{6}{c}{ \textit{Small building (50~m~x~50~m~x~20~m)}} \\ \hline

\begin{tabular}[c]{@{}c@{}}ColMap\end{tabular} & \textbf{0.11}      & 0.80      & \textbf{760} & \add{23084} & \add{23844} \\ 
\begin{tabular}[c]{@{}c@{}}\sysname \end{tabular}      & 0.12 & \textbf{0.09} & 864 &  \textit{\add{in-flight}} & \textbf{\add{864}}\\ \hline
\vspace{-0.0cm} &

\end{tabular}
}

\label{tab:e2e_reconstruction_quality_airsim}
\end{table}

\mradd{For both buildings, \sysname achieves: a) comparable, if not better, accuracy, b) an order of magnitude higher completeness, and c) 25x faster reconstruction as compared to ColMap.
}
\sysname achieves cm-level accuracy and completeness for both buildings (\tabref{tab:e2e_reconstruction_quality_airsim}). ColMap (like \sysname) depends on trajectory design. For ColMap, we tried multiple trajectories, including ones suggested for drone photogrammetry~\cite{dronedeploy}. From these, we report the trajectory with the best reconstruction quality. Without proper trajectory design, ColMap reconstruction fails altogether. With proper trajectory planning, ColMap reconstructs both buildings within 0.16~m accuracy and 0.80~m completeness. \sysname outperforms ColMap because it has real-time insight into the drone's tracking accuracy and can re-calibrate on-the-fly.

ColMap reconstructions are almost as accurate as \sysname but highly incomplete. This is because of the inherent nature of photogrammetry-based reconstructions; \textit{if ColMap cannot converge to a good solution for a given region, it will leave that region empty, \add{resulting in a model with many incomplete regions} (as shown in~\figref{fig:building_mvs}).} \add{As such, \sysname constructs 3D models with comparable if not better accuracy, and an order of magnitude higher completeness than those generated by ColMap.}

\add{
\sysname creates 3D models 25x faster than ColMap.
It builds 3D models for the small and large buildings in approximately 15~mins and 20~mins (inclusive of flight times), respectively.
Because it uses online reconstruction with cloud offload, \sysname simultaneously builds a 3D model at the cloud as it receives streamed point clouds from the drone.
This approach makes available the entire 3D model as soon as the model collection flight is complete (details in \secref{s:other-meas-sysn}) (as indicated in \tabref{tab:e2e_reconstruction_quality_airsim}).
For ColMap the end-to-end reconstruction time is the sum of the flight time and the processing time.
ColMap takes significantly more time to process the collected data because it uses compute-intensive photogrammetry-based reconstruction.
Even with a powerful GPU-equipped machine, ColMap takes approximately 6.5~hours and 9.2~hours to build 3D models of the small and large buildings, respectively{\footnote{We do not evaluate \sysname against smartphone-based LiDAR reconstruction approaches because they are both inaccurate and highly inefficient for outdoor reconstruction (\secref{sec:background_motivation}). Due to the smartphone's limited sensing range, model collection can take several hours and optimal trajectory planning takes well over 10 hours due to the large search space.}}.
}


\add{
The flight times for ColMap are relatively shorter (107~seconds shorter) than \sysname.
The planned trajectory length for \sysname's model collection was actually 185~seconds shorter than ColMap.
However, it incurred re-calibration flights, resulting in a longer flight time.
Moreover, its flight time also includes 150~seconds for reconnaissance.
Despite incurring only 107~seconds of additional flight time, \mradd{\sysname still reconstructs 25x faster than ColMap,} and improves accuracy and completeness by 7~cm and 70~cm respectively.
}

\subsubsection{Comparison: LiDAR SLAM-based Reconstruction and GPS}
Results from our real-world drone flights validate the \sysname outperforms LiDAR SLAM implementations~\cite{Cartographer, zhang2014loam} (\tabref{tab:e2e_reconstruction_quality_real}).
For this experiment, we compared \sysname against both scan and feature matching based SLAM implementations. 
We used Google Cartographer~\cite{Cartographer} and LOAM\add{~\cite{zhang2014loam}} as representatives for scan, and feature matching approaches, respectively.
We reconstructed the 3D model of a real-world 70~m~x~40~m~x~20~m rectangular building (\figref{fig:slam_vs_gps_3d_model}). Because we lack a reference ground truth for real-world data, we use the 3D model generated from raw, uncompressed traces.
\tabref{tab:e2e_reconstruction_quality_real} summarize the results.
3D reconstruction with Cartographer and LOAM fail completely for the same reasons mentioned above for ColMap (\ie no trajectory planning, and no re-calibration). 
With GPS, it is possible to do in-flight reconstruction, however, the accuracy and completeness being 1.60~m and 0.53~m, make such 3D models unusable. 
With \sysname, for this building, we can build accurate, and complete 3D models whose completeness and accuracy are 9~cm and 13~cm, respectively.
These experiment also demonstrates that \sysname performs equally well on real-world traces captured from our drone prototype.

\begin{table}[t]
\caption{Reconstruction quality of a \textit{real-world} 70~m~x~40~m~x~20~m building for \sysname, two LiDAR SLAM implementations and GPS-based reconstruction relative to an uncompressed trace.}
\small{
  \vspace{0.1cm}

\centering
\begin{tabular}{c c c c}
\hline \hline
\begin{tabular}[c]{@{}c@{}}\textbf{Scheme}\end{tabular} & 
\begin{tabular}[c]{@{}c@{}}\textbf{BW (Mbps)}\end{tabular} & 
\begin{tabular}[c]{@{}c@{}}\textbf{Accuracy (m)}\end{tabular} & 
\begin{tabular}[c]{@{}c@{}}\textbf{Completeness (m)}\end{tabular} \\ \hline 
\begin{tabular}[c]{@{}c@{}}Cartographer~\cite{Cartographer}\end{tabular}         & 480 & 2.30 & 1.30 \\ 
\begin{tabular}[c]{@{}c@{}}LOAM~\cite{zhang2014loam}\end{tabular}         & 480 & 5.75 & 2.75 \\ 
\begin{tabular}[c]{@{}c@{}}GPS-based reconstruction\end{tabular}        & 3.80 & 1.60 & 0.53 \\ 
\begin{tabular}[c]{@{}c@{}}\sysname\end{tabular}             & \textbf{3.80} & \textbf{0.13} & \textbf{0.09} \\ \hline
\end{tabular}
}

\label{tab:e2e_reconstruction_quality_real}
\end{table}

\begin{table}
\caption{\system 3D reconstruction times (recon (\secref{sec:recon}) and model collection) and quality for different structures at low compression.}
\small{
\centering
\begin{tabular}{c c c c}
\hline \hline
\begin{tabular}[c]{@{}c@{}}\textbf{Structure type}\end{tabular} & 
\begin{tabular}[c]{@{}c@{}}\textbf{Flight duration (s)}\end{tabular} & 
\begin{tabular}[c]{@{}c@{}}\textbf{Accuracy (m)}\end{tabular} & 
\begin{tabular}[c]{@{}c@{}}\textbf{Completeness (m)}\end{tabular} \\ \hline 
Tall rectangle & 1430 & 0.08 & 0.06 \\ 
Large rectangle  & 1145 & 0.09 & 0.05 \\ 
Pentagonal   & 1204 & 0.11 & 0.08 \\ 
Small rectangle  & 864 & 0.12 & 0.09 \\ 
Gabled roof  & 862 & 0.13 & 0.07 \\ 
Plus-shaped  & 1063 & 0.16 & 0.07 \\ \hline
\vspace{-0.0cm} &

\end{tabular}
}

\label{tab:e2e_reconstruction_vs_building_types}
\end{table}

\subsubsection{Generalization to Different Building Shapes}

\sysname is designed to reconstruct a variety of building types (\secref{sec:design}). It reconstructs all these buildings within a single drone battery cycle (25 to 30~mins) at cm-level accuracy and completeness (\tabref{tab:e2e_reconstruction_vs_building_types}). Flight durations for fairly large buildings (100~m~x~50~m~x~20~m large rectangle, pentagon and 50~m~x~50~m~x~40~m tall building) are relatively longer because they require multiple re-calibrations. Even for these, \sysname preserves cm-level accuracy and completeness.

\subsection{Performance}
\label{s:other-meas-sysn}

\subsubsection{Feasibility of Reconstruction over Cellular}

\begin{figure}[t]
    \centering
    \includegraphics[width=0.5\columnwidth]{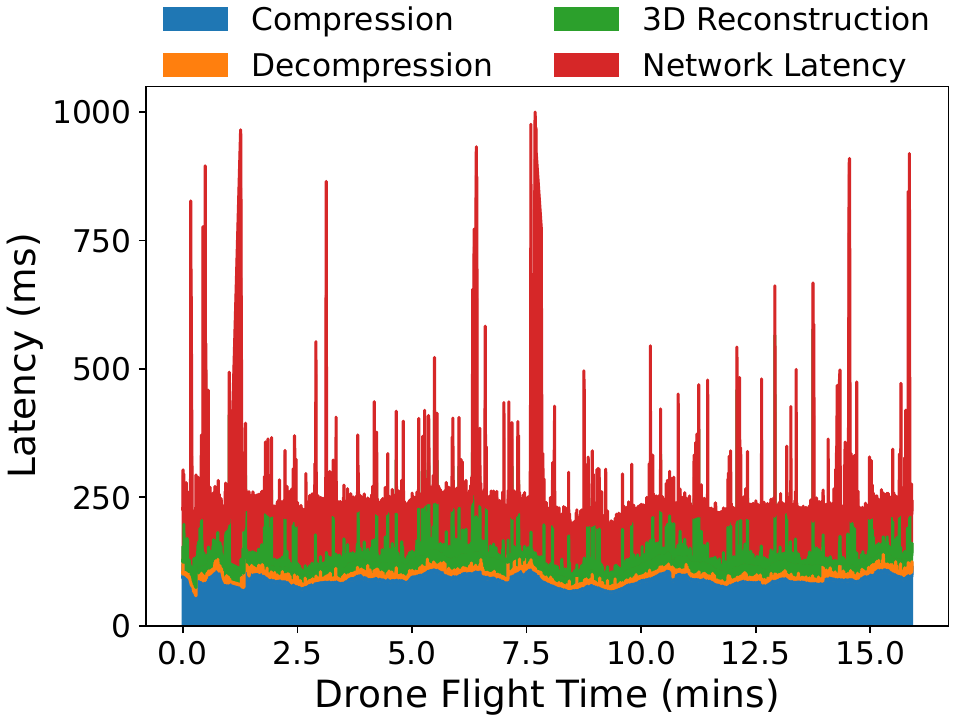}
    \caption{End-to-end latency from a real-world drone flight.
    \label{fig:e2e_latency}}
\end{figure}

To validate that \sysname can collect a 3D model in the real-world, we used our end-to-end implementation of \sysname to reconstruct a 70~m~x~40~m~x~20~m building by streaming compressed point clouds over LTE whilst the drone was in-flight (at 10~Hz) our AWS VM that ran 3D reconstruction. The experiment ran for about 16 minutes of drone flight (\figref{fig:e2e_latency} and \tabref{tab:e2e_tail}). \add{We are interested in two aspects of performance: end-to-end latency and the ability to process at frame rate.}

\sysname has a per-frame end-to-end 99th percentile latency of approximately half a second. This means that the cloud can detect excessive drift and initiate mid-flight recalibration within a second. An interesting side effect of this design is that the 3D model is ready within half a second of flight completion. The average end-to-end processing latency \textit{per frame} is about 234~ms. Of this, network latency accounts for nearly 104~ms. \add{This network latency is an artifact of our experimental setup: the drone, flying at a location on the west coast of the US streamed point clouds to a VM on the east coast. In practical settings, we expect offloads to nearby cloud regions, resulting in much lower end-to-end latency.} \add{To put these numbers in perspective, \sysname's latency is comparable to real-time 2D video conferencing systems (\eg Zoom, Skype), with a latency of 200-300~ms~\cite{videoconfimc, videotransfernsdi}, and is twice as fast as 360-degree video streaming services~\cite{poi360}. }

\add{\sysname can maintain LiDAR frame rate.} On-board point cloud compression takes on average 88~ms per frame, sufficient to sustain a 10~Hz LiDAR's frame rate. Decompression and 3D reconstruction are fairly fast on the cloud. Trajectory generation runs once and takes on average 14.6~ms per building. Drift estimation runs periodically and takes on average 11.2~ms.

\begin{table}[t]
\caption{\sysname enables fast 3D reconstruction over LTE. \add{Each row shows where the component in located (\ie on the drone or the cloud), its \textit{per frame} average latency, and \textit{per frame} 99th percentile latency.}}
\small{
  \vspace{0.1cm}

\begin{tabular}{c c c c }
\hline
\hline
\begin{tabular}[c]{@{}c@{}}\textbf{Component}\end{tabular} 
& \begin{tabular}[c]{@{}c@{}}\textbf{\add{Location}}\end{tabular}
& \begin{tabular}[c]{@{}c@{}}\textbf{Average (ms)} \end{tabular} 
& \begin{tabular}[c]{@{}c@{}}\textbf{99th percentile (ms)}\end{tabular} \\ \hline 

\begin{tabular}[c]{@{}c@{}}Compression\end{tabular} & \add{Drone} & 88.2 & 109.45  \\ 
\begin{tabular}[c]{@{}c@{}}Network latency\end{tabular} & \add{-} & 104.0 & 409.9  \\ 
\begin{tabular}[c]{@{}c@{}}Decompression\end{tabular} & \add{Cloud} & 8.9 & 11.7  \\ 
\begin{tabular}[c]{@{}c@{}}Reconstruction\end{tabular} & \add{Cloud} & 33.3 & 87.8  \\ 
\begin{tabular}[c]{@{}c@{}}End-to-end latency\end{tabular} & \add{-} & 234.0 & 552.1  \\ \hline
\end{tabular}
}

\label{tab:e2e_tail}
\end{table}

\begin{table}
\caption{Per-frame processing times for \sysname's building geometry estimation.}
\small{
  \vspace{0.1cm}

\begin{tabular}{c c c}
\hline \hline
\textbf{Stage} & 
\textbf{Sub-component} & 
{\begin{tabular}[c]{@{}c@{}}\textbf{Time (ms)}\end{tabular}} \\ \hline
\multirow{5}{*}{\begin{tabular}[c]{@{}c@{}}Surface extraction\end{tabular}} 
  & Point cloud decompression & 3.0 $\pm 0.3$ \\ 
& Surface normal estimation & 76.0 $\pm 82$ \\ 
& RANSAC plane-fitting & 5.0 $\pm 9.0$ \\ 
& Outlier removal & 0.2 $\pm 0.3$ \\ 
& Rooftop extraction & 6.0 $\pm 5.0$ \\ \cline {2-3} \hline
\multirow{1}{*}{\begin{tabular}[c]{@{}c@{}}Boundary estimation\end{tabular}}
& Rooftop stitching & 3.0 $\pm 2.0$ \\ \cline {2-3} \hline
 \multicolumn{2}{c}{Total time} & 93 $\pm 90.0$ \\ \hline
\end{tabular}
}

\label{tab:recon_performance}
\end{table}

\subsubsection{Rooftop Geometry Extraction Performance}

We profiled the execution time of each component in \sysname's building geometry extraction on a 15-minute \textit{real-world} trace. Point cloud compression executes on the drone, and other components run on our AWS VM. Extracting the building geometry requires 93~ms per frame (\tabref{tab:recon_performance}); with these numbers, we can sustain 10~fps. At this frame rate, our building detector is quite accurate (\secref{sec:recon}). The most expensive component is surface normal extraction (76~ms), even though we offload it to a GPU.
Thus, a moderately provisioned, cloud VM suffices to run \sysname at full frame rate with an end-to-end compute latency of about 100~ms for reconnaissance, and 33~ms for model collection. 



\subsection{Ablation and Sensitivity}
\label{s:ablation-study}


\begin{table}[t]
\centering
\caption{\add{Reconstruction accuracy, completeness, flight time, processing time, and end-to-end reconstruction times for two buildings using: a) \sysname without cloud offload (Offline-\sysname), and b) \sysname.}\fawad{Added table for cloud offload.}}
\small{
  \vspace{0.1cm}

\begin{tabular}{c c c c c c}
\hline \hline
\textbf{\add{Scheme}} & \add{\textbf{Acc. (m)}} & \add{\textbf{Compl. (m)}} & \add{\textbf{Flight (s)}} & \add{\textbf{Processing (s)}} & \add{\textbf{End-to-end reconstruction (s)}} \\ \hline

\multicolumn{6}{c}{ \add{\textit{Large building (100m~x~50m~x~20m)}}} \\ \hline

\begin{tabular}[c]{@{}c@{}}\add{Offline-\sysname}\end{tabular} & \add{0.39} & \add{0.27} & \textbf{\add{990}} & \add{5330} & \add{6320}  \\ 


\begin{tabular}[c]{@{}c@{}}\add{\sysname} \end{tabular} 
& \add{\textbf{0.09}} & \add{\textbf{0.05}} & \add{1430} & \add{\textit{in-flight}} & \add{\textbf{1430}} \\ \hline

\multicolumn{6}{c}{ \add{\textit{Small building (50m~x~50m~x~20m)}}} \\ \hline

\begin{tabular}[c]{@{}c@{}}\add{Offline-\sysname}\end{tabular} & \add{0.29} & \add{0.15} & \textbf{\add{720}} & \mradd{4395} & \add{5115} \\

\begin{tabular}[c]{@{}c@{}}\add{\sysname} \end{tabular} 
& \add{\textbf{0.12}} & \add{\textbf{0.09}} & \add{864} & \add{\textit{in-flight}} & \add{\textbf{864}} \\ \hline

\end{tabular}
}

\label{tab:cloud_offload}
\end{table}

\subsubsection{\add{Cloud Offload}}

\add{To show the importance of cloud offload, we compared \sysname to an offline approach that does not use cloud offload (Offline-\sysname in \tabref{tab:cloud_offload}). Offline-\sysname uses \sysname's trajectory planning, then collects and stores \textit{raw} 3D point clouds on the drone, and processes them offline after the drone has landed. Its accuracy is 3x worse, and completeness 3.5x worse (on average) as compared to a complete \sysname implementation with cloud offload. Without cloud offload, offline-\sysname cannot track and mitigate drift errors in real-time. The flight times for offline-\sysname are relatively shorter because there are no re-calibration flights. However, \sysname's end-to-end reconstruction time is 6x smaller than offline-\sysname because it pipelines model collection and 3D reconstruction, whereas the latter first collects and then processes point clouds into a 3D model.}


\subsubsection{Trajectory Planning}

We evaluated \sysname without its optimized trajectory generation. Instead, we use a rectilinear trajectory (\figref{fig:rectflight}) that satisfies the minimum point density requirement (\sysname-st, \tabref{tab:trajectory_planning}). \sysname-st has poorer accuracy and completeness relative to \sysname, even though it has a longer duration flight (which should, in theory, give it a better chance to capture more detail of the building). However, a larger proportion of \sysname-st's trajectory is not in a parallel orientation; this illustrates the importance of explicitly optimizing for parallel flights in the objective function (\eqnref{eq:opt}, in \secref{s:slam-phase}).

\begin{table}[t]
\centering
\caption{\add{Reconstruction accuracy, completeness, and flight time for two buildings using: a) \sysname with a simple rectilinear trajectory (\sysname-st), and b) \sysname.}}
\small{
  \vspace{0.1cm}

\begin{tabular}{c c c c}
\hline
\hline
\textbf{Scheme} & \textbf{Accuracy (m)} & \textbf{Completeness (m)} & \textbf{Flight time (s)} \\ \hline
\multicolumn{4}{c}{ \textit{Large building (100~m~x~50~m~x~20~m)}} \\ \hline

\begin{tabular}[c]{@{}c@{}}\sysname-st\end{tabular}        & 0.21 & 0.24 & 1520 \\ 

\begin{tabular}[c]{@{}c@{}}\sysname \end{tabular} 
& \textbf{0.09} & \textbf{0.05} & \textbf{1430} \\ \hline

\multicolumn{4}{c}{ \textit{Small building (50~m~x~50~m~x~20~m)}} \\ \hline

\begin{tabular}[c]{@{}c@{}}\sysname-st\end{tabular}        & 0.25 & 0.14 & 900  \\ 

\begin{tabular}[c]{@{}c@{}}\sysname \end{tabular} 
& \textbf{0.12} & \textbf{0.09} & \textbf{864} \\ \hline
\vspace{-0.1cm}&

\end{tabular}
}

\label{tab:trajectory_planning}
\end{table}

\subsubsection{Re-calibration}

To show the effect of in-flight re-calibration, we compare reconstruction quality with (w) and without (w/o) recalibration in AirSim (\tabref{tab:reconstruction_with_recalibration}). Across our six buildings, on average, at the expense of 35\% (285~seconds) longer flights, \sysname improves accuracy by 70\% (28~cm) and completeness by 64\% (13~cm) with re-calibration flights. Larger buildings (tall rectangle, large rectangle, and pentagonal) require longer aerial flights which accumulate higher drift. This results in relatively more re-calibration flights and hence higher flight duration. \add{Apart from these large buildings which required two re-calibration flights, the remaining buildings require a single re-calibration flight.} Even so, \sysname is able to reconstruct these buildings accurately (within cm-level accuracy and completeness), demonstrating the importance of re-calibration whilst in-flight.

\subsubsection{Sensitivity: Compression Levels}

We explore the impact of compression on accuracy and completeness using (a) a synthetic building in AirSim and (b) \textit{real-world} traces. In addition to the three compression schemes discussed earlier (\secref{s:compression}), we compute accuracy and completeness for (a) viewpoint compression and (b) lossless compression. The first contextualizes our results, while the second alternative explores reconstruction performance under higher bandwidth as would be available, \eg in 5G deployments.


\begin{table}[t]
\captionsetup[subtable]{labelfont=rm}
\caption{Ablation study results.}
\begin{subtable}{\columnwidth}
\centering
\small{
\caption{Flight duration and reconstruction quality for buildings at low compression with (w) and without (w/o) re-calibration.\label{tab:reconstruction_with_recalibration}
}
\begin{tabular}{c c c c c c c}
\hline \hline
\multirow{2}{*}{\begin{tabular}[c]{@{}c@{}}\textbf{Structure} \\\textbf{type}\end{tabular}} & 
\multicolumn{2}{c }{\textbf{Flight duration (s)}} & 
\multicolumn{2}{c }{\textbf{Accuracy (m)}} & 
\multicolumn{2}{c}{\textbf{Completeness (m)}} \\ 
& 
\begin{tabular}[c]{@{}c@{}}\textit{w/o}\end{tabular} & 
\begin{tabular}[c]{@{}c@{}}\textit{w}\end{tabular} & 
\begin{tabular}[c]{@{}c@{}}\textit{w/o}\end{tabular} & 
\begin{tabular}[c]{@{}c@{}}\textit{w}\end{tabular} & 
\begin{tabular}[c]{@{}c@{}}\textit{w/o}\end{tabular} & 
\begin{tabular}[c]{@{}c@{}}\textit{w}\end{tabular} \\ \hline 

Tall rect. & 1147 & 1430 & 0.31 &  \textbf{0.08} & 0.15 & \textbf{0.06} \\ 
Large rect.   & 855 & 1145 & 0.87 & \textbf{0.09} & 0.35 & \textbf{0.05} \\ 
Pentagonal    & 891 & 1204 & 0.61 & \textbf{0.11} & 0.22  & \textbf{0.08} \\ 
Small rect.  & 678 & 864 & 0.53 & \textbf{0.12} & 0.13 & \textbf{0.09} \\ 
Gabled roof  & 667 & 862 & 0.32 & \textbf{0.13} & 0.11 &  \textbf{0.07} \\ 
Plus-shaped  & 620 & 1063 & 0.48 & \textbf{0.16} & 0.51  & \textbf{0.07} \\ \hline
\vspace{-0.1cm}&

\end{tabular}
}

\end{subtable}
\begin{subtable}{\columnwidth}
\centering
\small{
\caption{The impact of compression on accuracy and/or completeness.\label{tab:reconstruction_vs_compression}} 
\begin{tabular}{c c c c}
\hline \hline
\begin{tabular}[c]{@{}c@{}}\textbf{Compression profile}\end{tabular} & 
\begin{tabular}[c]{@{}c@{}}\textbf{BW (Mbps)}\end{tabular} & 
\begin{tabular}[c]{@{}c@{}}\textbf{Accuracy (m)}\end{tabular} & 
\begin{tabular}[c]{@{}c@{}}\textbf{Completeness (m)}\end{tabular} \\ \hline 
\multicolumn{4}{c}{\textit{\textbf{Real-world} 70~m~x~40~m~x~20~m large building}} \\ \hline
\begin{tabular}[c]{@{}c@{}}View-point\end{tabular}  & 42.7  & 0.00 & 0.00 \\ 
\begin{tabular}[c]{@{}c@{}}Lossless\end{tabular}    & 7.86  & 0.06 & 0.07 \\ 
\begin{tabular}[c]{@{}c@{}}Low\end{tabular}         & 3.80  & 0.13 & 0.09 \\ 
\begin{tabular}[c]{@{}c@{}}Medium\end{tabular}      & 2.50  & 0.23 & 0.16 \\ 
\begin{tabular}[c]{@{}c@{}}High\end{tabular}        & 1.27  & 0.28 & 0.29 \\ \hline

 \multicolumn{4}{c}{\textit{\textbf{AirSim} 50~m~x~50~m~x~20~m small building}} \\ \hline
\begin{tabular}[c]{@{}c@{}}View-point\end{tabular}  & 42.7  & 0.10 & 0.08 \\ 
\begin{tabular}[c]{@{}c@{}}Lossless\end{tabular}    & 7.86  & 0.10 & 0.10 \\ 
\begin{tabular}[c]{@{}c@{}}Low\end{tabular}         & 3.80  & 0.12 & 0.09 \\ 
\begin{tabular}[c]{@{}c@{}}Medium\end{tabular}      & 2.50  & 0.35 & 0.09 \\ 
\begin{tabular}[c]{@{}c@{}}High\end{tabular}        & 1.27  & 0.44 & 0.10 \\ \hline
\vspace{-0.0cm} &
\end{tabular}
}


\end{subtable}

\label{tab:ablation}
\end{table}


Viewpoint filtering is the same as a raw point cloud, but with zero points removed. With this, it achieves a 10$\times$ compression throughout. As \tabref{tab:reconstruction_vs_compression} shows, low compression is an order of magnitude more efficient beyond this. Despite this, \sysname can achieve high quality reconstruction. For the AirSim building, consider accuracy: the viewpoint filtered point cloud has an accuracy and completeness of 0.10~m and 0.08~m respectively, which is attributable entirely to SLAM error. Low compression, with a bandwidth of 3.8~Mbps (easily achievable over LTE and over 100$\times$ more compact than the raw LiDAR output) only adds 2~cm and 1~cm to accuracy and completeness, respectively. This also shows that there is little room for improvement with 5G bandwidths. Medium and high compression have significantly poorer accuracy and completeness. Results for other buildings are similar, so we omit them for brevity.
\raj{Some parts of the paper have accuracy and completeness numbers in cm, while others have in meters, particularly the tables.}

Results from our drone flights validate that \textit{real-world} data of a large building (dimensions in \tabref{tab:reconstruction_vs_compression}) results in comparable performance (\tabref{tab:reconstruction_vs_compression}). Since we lack a reference ground truth for real-world data, we use the 3D model generated from raw traces. With real-world traces, we can build accurate, and complete 3D models that are within 9-13~cm completeness and accuracy for low compression, and about 16-23~cm for medium compression, relative to the raw traces. This suggests that highly compressed point clouds do not significantly impact accuracy and completeness.

\begin{figure*}[t]
  \centering
     \includegraphics[width=0.95\textwidth]{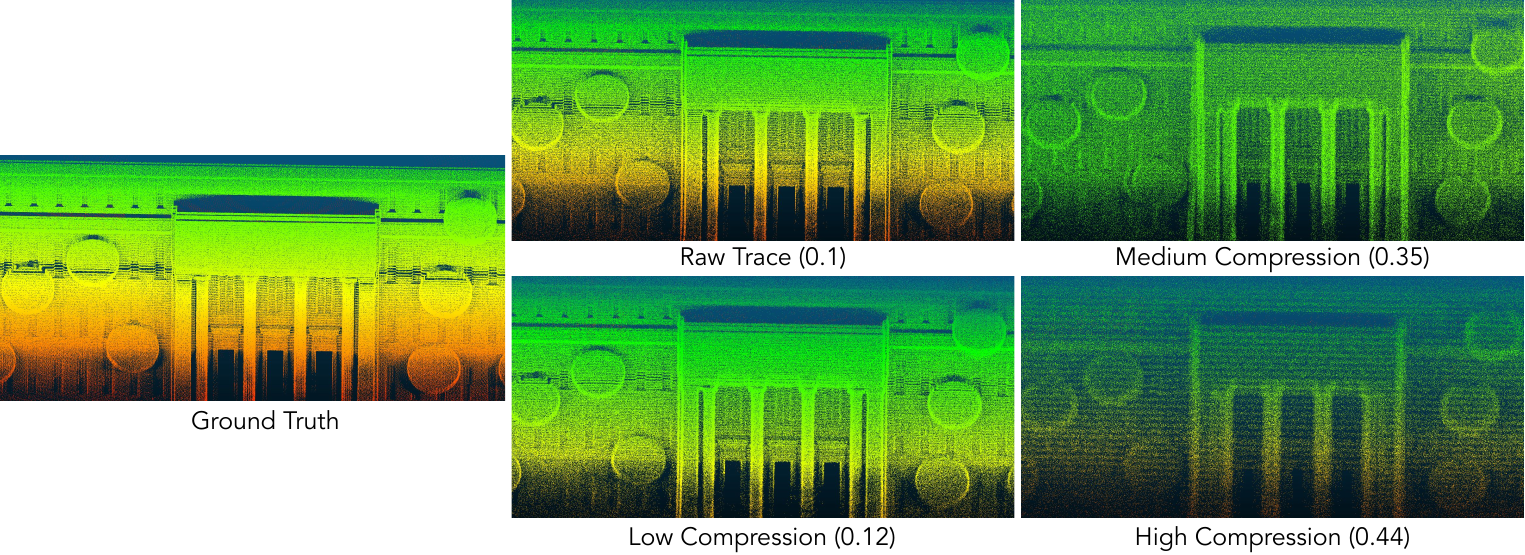}
  \caption{3D models accuracy at different compression levels.}
  \label{fig:3d_model}
\end{figure*}

To get a visual feel for the degradation resulting from lower accuracy,  \figref{fig:3d_model} shows the ground-truth model, together with the \sysname reconstructions at different compression levels. With an accuracy of 0.12~m (with 3.8~Mbps upload bandwidth), the model closely matches the ground truth. As accuracy worsens at higher compression levels, the textures on the building increasingly become less distinct and start to show some small artifacts, arising not because of compression but because of SLAM imperfections (\secref{s:ablation-study}).

\begin{table}[t]
\centering
\caption{\add{Reconstruction accuracy, completeness, and flight time for \sysname with different flight speeds.}}
\vspace{0.1cm}
\small{
\begin{tabular}{c c c c}
\hline
\hline
\add{
\textbf{Speed (m/s)}} & \add{\textbf{Accuracy (m)}} & \add{\textbf{Completeness (m)}} & \add{\textbf{Flight time (s)}} \\ \hline

\begin{tabular}[c]{@{}c@{}}\add{0.5}\end{tabular}        & \add{0.43} & \add{0.34} & \add{2045} \\ 

\begin{tabular}[c]{@{}c@{}}\add{1} \end{tabular} 
& \add{\textbf{0.09}} & \add{\textbf{0.05}} & \add{1145} \\


\begin{tabular}[c]{@{}c@{}}\add{2}\end{tabular}        & \add{0.91} & \add{0.45} & \add{\textbf{768}}  \\ \hline
\vspace{-0.1cm}&


\end{tabular}
}

\label{tab:flight_speeds}
\end{table}

\subsubsection{Sensitivity: Target Density}

We evaluated \sysname at two different densities: 7.5 points per m$^2$ and 1 point per m$^2$. The lower density flight took only 31\% of the higher density flight time. However, as expected, reconstruction is worse at lower target densities (0.18~m accuracy, 0.24~m completeness vs. 0.08~m and 0.06~m for the higher density). For applications that can tolerate this degradation, the smaller flight time might result in cost savings.

\subsection{Data Collection Parameter Choices}
\label{s:data-collection}

\sysname determines model collection flight parameters with a parameter sweep in simulation and on \textit{real-world traces}  (\secref{s:slam-phase}). In simulations, we evaluated SLAM error for every combination of drone speed (0.5~m/s to 3~m/s), distance from building (10~m to 40~m), and LiDAR orientation (ranging from parallel to perpendicular). 

\subsubsection{Orientation}

\figref{fig:parallel_vs_other_orientations} plots SLAM error as a function of LiDAR orientation (\figref{fig:lidar_orientation}) with respect to the direction of motion. A parallel orientation has the lowest SLAM error (in \figref{fig:parallel_vs_other_orientations}, yaw 0$\degree$ is parallel and yaw 90$\degree$ is perpendicular), because it has the highest overlap between successive frames; as yaw increases, overlap decreases, resulting in higher SLAM error (\secref{s:slam-phase}). 

\subsubsection{Distance}

\figref{fig:parallel_slam_vs_height_speed} plots the SLAM error as a function of the drone's distance from the building surface for the \textit{parallel} orientation of the LiDAR. Error increases slowly with height; beyond a 20~m distance from the building, the error is more than 1~m. Point densities decrease with height and affect SLAM's ability to track features/points across frames (\secref{s:slam-phase}). Flying close to the building surface reduces scan width and hence increases flight duration. Rather than fly lower, \sysname operates at a 20~m distance from the building to reduce flight duration.



\subsubsection{Speed}

\begin{figure}[t]
\captionsetup[subfigure]{labelfont=rm}
\begin{subfigure}{0.35\columnwidth}
    \centering
 \includegraphics{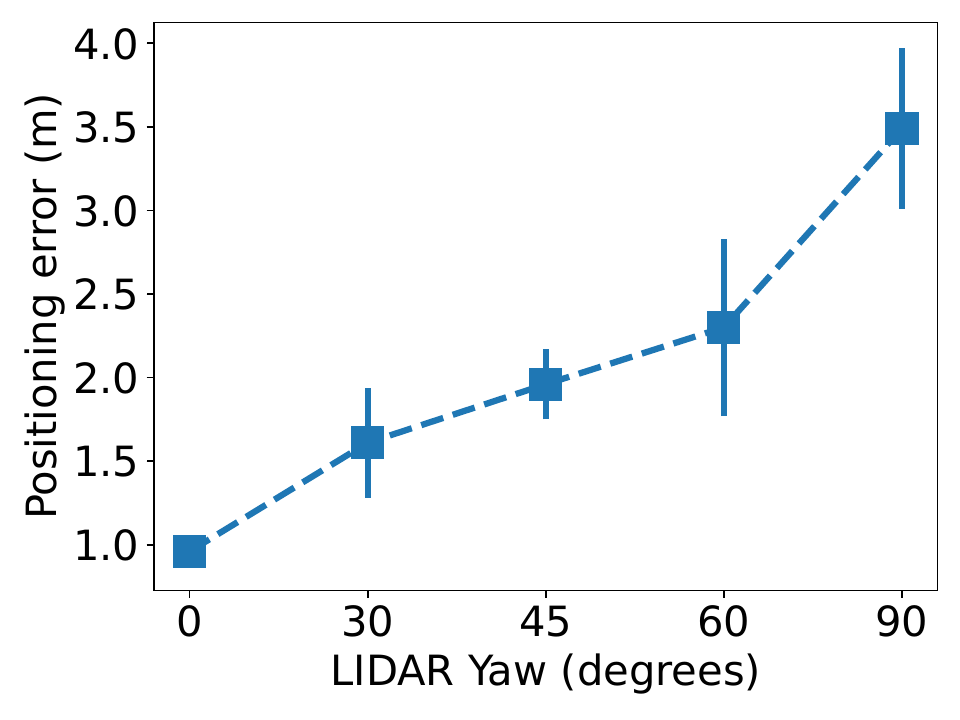}
    \caption{SLAM errors for various LiDAR orientations.}
    \label{fig:parallel_vs_other_orientations}
\end{subfigure}
\hspace{0.5cm}
\begin{subfigure}{0.35\columnwidth}
\centering
 \includegraphics{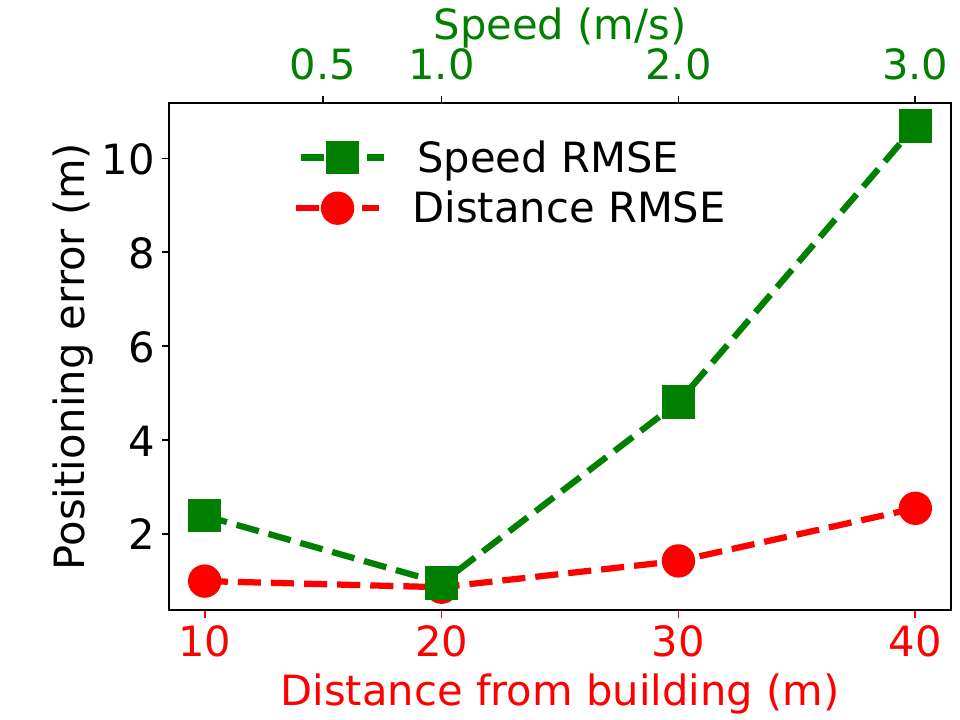}
    \caption{SLAM error for speeds and building distances.}
    \label{fig:parallel_slam_vs_height_speed}
\end{subfigure}
\caption{Parameter study for optimal flight parameters in model collection.}
\label{fig:param_study}
\end{figure}

Speed impacts SLAM positioning significantly (\figref{fig:parallel_slam_vs_height_speed}). Beyond 1~m/s, SLAM cannot track frames accurately due to lower overlap between frames (\secref{s:slam-phase}). Below 1~m/s \ie at 0.5~m/s, the flight duration (in seconds) is twice that of 1~m/s resulting in drift accumulation. For accurate reconstruction, \sysname flies the drone at 1~m/s. \add{\tabref{tab:flight_speeds} explains the impact of drone speed on reconstruction quality. Although flying faster \ie at 2~m/s decreases the flight time by 33\%, this comes at the cost of 80~cm in accuracy and 40~cm in completeness relative to flying at 1~m/s. This is because at higher speeds, SLAM cannot accurately track frames because of lower overlap. On the other hand, flying slower \ie at 0.5~m/s doubles the flight duration which causes SLAM to accumulate more drift, hence adding 34~cm in accuracy and 29~cm in completeness. \sysname flies the drone at 1~m/s which minimizes accuracy, completeness, and flight time.}





\subsubsection{Real-world Validation}

To validate these observations, we performed a parameter sensitivity study on real-world flights to determine the optimum parameters for SLAM positioning. 
For the lack of accurate ground truth in the real-world, we compare SLAM positions against a GPS trace. Because GPS is erroneous, we only draw qualitative conclusions. 


As \tabref{tab:real_speed_height_vs_rmse}, taken from our drone traces, shows, slower flights have lower SLAM error than faster ones, and parallel orientations have lower SLAM error than perpendicular. Similarly, SLAM error increases with height and, in real-world traces, the parallel orientation seems to be significantly better than the perpendicular orientation (\tabref{tab:real_speed_height_vs_rmse}). At a distance of 20~m from the surface of the building, the parallel orientation has the minimum positioning error \ie 1.25~m. Beyond 40~m for parallel and 20~m for perpendicular, SLAM loses track completely because of lower point density.


\begin{table}
\caption{Positioning errors for parallel and perpendicular orientations at different speeds (at distance 20~m) and errors for different distances (at a speed of 1~m/s) for \textit{real-world traces}.}
\centering
\vspace{0.1cm}
\small{
\begin{tabular}{ccccc}
\hline \hline
\multirow{2}{*}{\begin{tabular}[c]{@{}c@{}}\textbf{LiDAR}\\ \textbf{Orientation}\end{tabular}}
& \multicolumn{2}{c}{\textbf{Speed (m/s)}} & 
\multicolumn{2}{c}{\textbf{Distance (m)}} \\ 
              & \textit{1.5} & \textit{3.0} & \textit{20}  & \textit{40} \\  
              \hline
Parallel      & 1.3 & 3.3 & 1.2  & 5.4 \\ 
Perpendicular & 3.1 & 7.6 & 2.1  & $\infty$ \\ \hline
\vspace{-0.1cm}
\end{tabular}
}

\label{tab:real_speed_height_vs_rmse}
\end{table}

\subsection{Rooftop Geometry}
\label{sec:eval:building}

We use two metrics for rooftop geometry estimation: accuracy, and completeness. Accuracy is the average (2D) distance between each point (quantized to 0.1~m) on the predicted rooftop boundary and the nearest point on the actual building boundary. Completeness, is the average distance between each point on the actual boundary and the nearest point on \sysname's predicted boundary. Lower values of accuracy and completeness are better. We use real-world traces collected from the \system prototype and synthetic traces from AirSim. For real-world traces, we pinpointed the building's boundary on Google Maps~\cite{google_maps} for ground truth. For AirSim, we collected the ground truth from the Unreal Engine.

\begin{figure}[t]
\captionsetup[subfigure]{labelfont=rm}
\begin{subfigure}{0.35\columnwidth}
    \centering
 \includegraphics{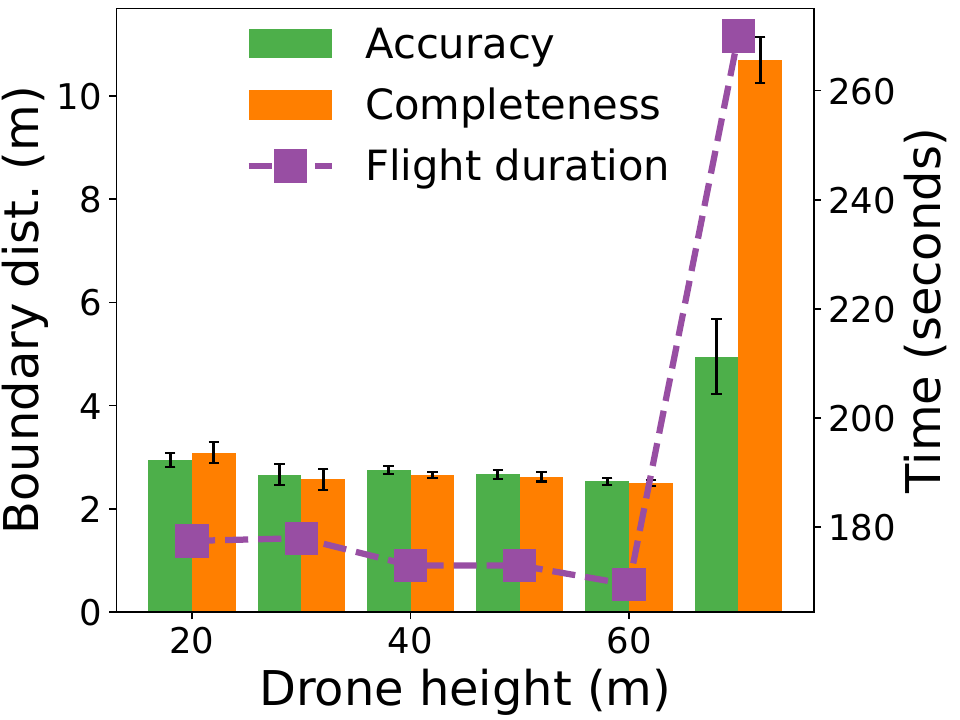}
    \caption{A recon flight height of 60~m above the building surface minimizes accuracy, completeness (left y-axis) and flight duration (right y-axis).}
    \label{fig:recon_height}
\end{subfigure}
\hspace{0.5cm}
\begin{subfigure}{0.35\columnwidth}
\centering
 \includegraphics{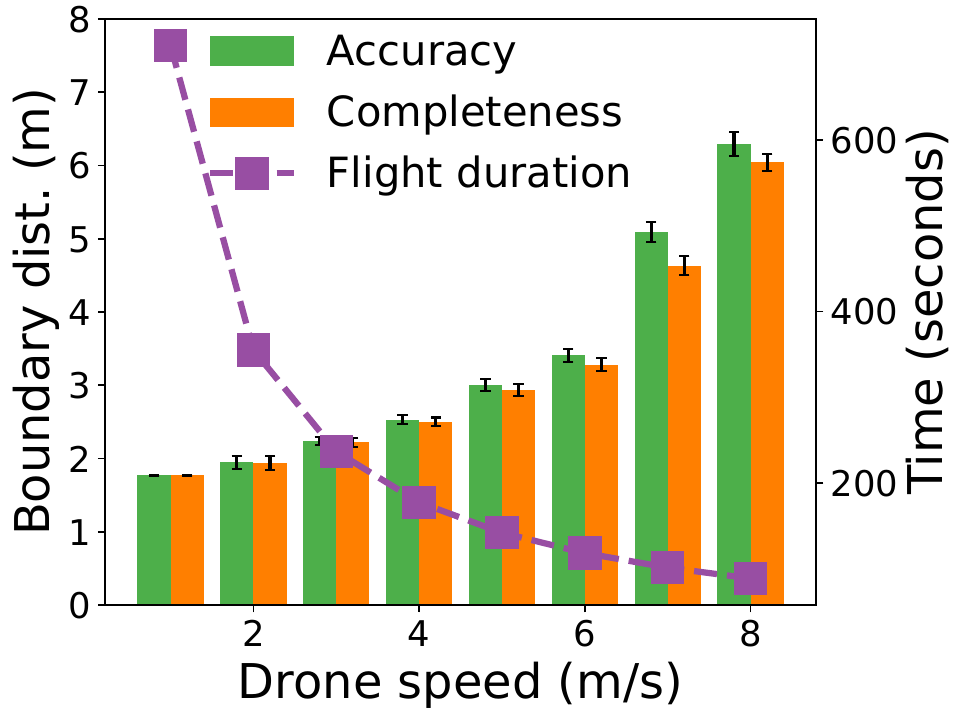}
    \caption{A recon flight speed 4-5~m/s ensures low accuracy, completeness (left y-axis) and flight duration (right y-axis).}
    \label{fig:recon_speed}
\end{subfigure}
\caption{Rooftop geometry extraction accuracy, completeness, and flight duration as function of drone height, and speed.}
\label{fig:recon_eval}
\end{figure}

\tabref{tab:recon_performance} shows the execution time for rooftop geometry estimation, on \textit{real-world} traces, with the cloud VM. The average processing time per frame is 93~ms, dominated by GPU-accelerated surface normal estimation (76~ms). This can sustain 10~fps.




\sysname also extracts rooftop geometry accurately. Across 3 \textit{real-world traces} collected over a 70m~x~60m~x~20m building, its average accuracy is 1.42~m and completeness is 1.25~m, even at the highest compression and when it samples every other frame. These results justify our choice of a fast, high, reconnaissance flight (\secref{sec:recon}).

We extensively evaluated robustness to different building shapes (\tabref{tab:e2e_reconstruction_vs_building_types}), point cloud compression (\tabref{tab:reconstruction_vs_compression}), subsampling, and flight parameters (\figref{fig:recon_height} and \figref{fig:recon_speed} justify the choice of the recon speed and height). We omit the details for brevity but report the following: (a) recon flights can be short (boundary detection is insensitive to point density and overlap), so it can use perpendicular orientation to reduce flight duration, fly at 60~m from the building's surface (\figref{fig:recon_height}) at 4~m/s (\figref{fig:recon_speed}); (b) it tolerates subsampling up to 1~Hz; (c) it generalizes to different building geometry described in \tabref{tab:roof_types}.

\section{Discussion and Future Work}
\label{sec:discussion}

 
\subsubsection*{\mradd{Generalization to Other Buildings and Structures}}

In this paper, we have focused on reconstructing large outdoor regular-shaped buildings, which make up more than 99\% of residential and commercial buildings in large cities (\tabref{tab:roof_types}). \sysname cannot reconstruct the remaining 24 buildings in \tabref{tab:roof_types} because its rooftop geometry extractor and 3D unfolding algorithms were not designed for buildings with hemispherical (domes like cathedrals and mosques), and ellipsoidal roofs (the Empire State Building). We have left extending \sysname to support these buildings and other large structures like bridges and stadiums to future work. 
\add{As an aside, for large buildings with model collection flights that exceed the drone's battery time, \sysname can return and land the drone at the origin (similar to a re-calibration maneuver), swap the battery, and then resume the remaining model collection flight.}

\add{\sysname's model collection techniques are applicable to other large physical structures such as blimps and aircraft. We encapsulated a blimp in a 3D bounding box, and reconstructed it using \sysname with an accuracy of 20~cm and completeness of 3~cm. Further examination and extension of \sysname to such structures is left for future work.}

\subsubsection*{\mradd{Reducing End-to-end Latency}}

\add{
\sysname's average end-to-end latency is 234~ms, with compression and network latency contributing 88~ms and 104~ms, respectively. Recent work~\cite{paralleloctree} has demonstrated 1-2 orders of magnitude improvements in octree search, which can lead to faster compression times.
Offload to edge compute instead of a remote cloud instance can reduce network latency to \mradd{20ms or less~\cite{edge_offload_3, edge_offload_4}.
}}

\subsubsection*{\mradd{Cloud Offload}}

In the future, mobile compute capabilities will improve. However, at the same time, LiDAR technology continues to evolve and 128 beam LIDARs are already available, and these will require significantly more compute. So it is unclear if, or when, accurate LiDAR SLAM can run entirely on the drone. Even if that becomes feasible, \sysname's trajectory optimization and re-calibration algorithms will be relevant for accurate 3D modeling.

\subsubsection*{\mradd{Mobile Device LiDAR-based Reconstruction}}

\sysname could have used LiDARs now available on mobile devices (\eg iPhones) to perform 3D reconstruction. However, LiDARs on mobile devices work well only in indoor environments because: a) the mobile device LiDAR has a limited sensing range of about 3~m~\cite{iphone_depth}, and b) has high depth estimation errors beyond this range \eg up to 30~cm at 4~m~\cite{iphone_depth2}.  In outdoor spaces, these approaches can have tracking errors in meters~\cite{iphone_tracking} which leads to undesirable results.

\subsubsection*{\mradd{Infrastructure Support to Improve \sysname}}

Loop closure is a critical component in SLAM and \sysname to reduce drift error. \sysname invokes loop closure when it detects excess drift; this leads to longer flight times. For future work, \sysname can plant April Tags~\cite{mateos2020apriltags} at well-defined locations on the buildings to recalibrate SLAM (hence invoking loop closure) without the drone having to fly back to the origin. Finally, \sysname uses flight distance as a proxy for drone battery. Future work can explore incorporating more sophisticated battery models into the trajectory optimization.

\section{Related Work}
\label{sec:related}

\subsubsection*{\mradd{Networked 3D Sensing}}

Recent work has explored, in the context of cooperative perception~\cite{AVR,emp,vieye,shi2022vips} and real-time 3D map updates~\cite{CarMap}, transmitting 3D sensor information over wireless networks. Compared to \sysname, they use different techniques to overcome wireless capacity constraints. 
\mradd{
Of these, VI-Eye~\cite{vieye} and VIPS~\cite{shi2022vips} propose registration techniques to align a pair of point clouds (one captured from a vehicle, and another from an infrastructure-mounted LiDAR).
To do this, they use additional scene information (\eg road landmarks, and vehicle bounding boxes).
As with ICP, progressively using VI-Eye and VIPS to align large number of point clouds can result in significant drift.
In contrast, \sysname uses drift detection, and mitigation techniques (in addition to SLAM's bundle adjustment), making it more suitable for aligning thousands of raw point clouds accumulated over large drone flights.
}

\subsubsection*{\mradd{Drone Positioning}}
The robotics literature has studied efficient coverage path-planning for single~\cite{sensorplanning}, and multiple drones\add{~\cite{ilp-karthik}}. \sysname's trajectory design is influenced by more intricate constraints like SLAM accuracy and equi-density goals. Accurately inferring drone motion is important for SLAM-based positioning~\cite{observability}. Cartographer~\cite{Cartographer}, which \sysname uses for positioning, utilizes motion models and on-board IMUs for estimating motion. A future version of \sysname can leverage drone orchestration~\cite{beecluster} and SLAM edge-offloading~\cite{edgeslam} for larger scale reconstruction. 

\subsubsection*{\mradd{Offline Reconstruction using Images}}
UAV photogrammetry~\cite{federman2017} reconstructs 3D models offline from 2D images. Several pieces of work~\cite{7139681, 7989530, 8124461, 8628990} study the use of RGB, and RGB-D cameras on UAVs for 3D reconstruction. Prior work~\cite{7139681} has proposed a real-time, interactive interface into the reconstruction process for a human guide. The most relevant of these~\cite{mostegel2016uav, 7422384} predicts the completeness of 3D reconstruction in-flight, using a quality confidence predictor trained offline, for a better offline 3D reconstruction. However, unlike \sysname, this work requires human intervention, computes the 3D model offline, requires close-up flights, cannot ensure equi-dense reconstructions, cannot dynamically re-calibrate for drift and is not an end-to-end system. A body of work has explored factors affecting reconstruction accuracy: sensor error~\cite{6899451}, tracking drift, and the degree of image overlap~\cite{7139681, LIENARD2016264}. Other work~\cite{8793729, 8628990, bylow2019combining} has explored techniques to reduce errors by fusing with depth information, or using image manipulations such as upscaling. Unlike \sysname, almost all of this work reconstructs the 3-D model offline.

\subsubsection*{\mradd{Offline Reconstruction Using LiDAR or Radar}}
3D model reconstruction using LiDAR~\cite{uav_lidar_1,uav_lidar_2} relies on additional positioning infrastructure such as base stations for real-time kinematic (RTK) positioning, and long-range specialized LiDAR to achieve tens of centimeters model accuracy. \sysname explores a different part of the design space: online reconstruction with sub-meter accuracy using commodity drones, GPS and LiDAR. Recent work has explored drone-mounted LiDAR based offline reconstruction of tunnels and mines, but require specialized LiDARs and a human-in-the-loop~\cite{Prometheus, hovermap} for drone guidance (either manually or by defining a set of waypoints). Finally, mmMesh~\cite{xue2021mmmesh} explores cloud-offload for human mesh reconstructions using millimeter-wave radar.

\section{Conclusions}
\label{s:conclusions}

In this paper, we have taken a step towards accurate, near-real time 3D reconstruction using drones. Our system, \sysname, uses novel techniques for navigating the tension between cellular bandwidths, SLAM positioning errors, and compute constraints on the drone. It contains algorithms for optimized trajectory generation, and for determining excessive SLAM drift. It can achieve reconstruction accuracy to within 10~cm in near real-time, even after compressing LiDAR data enough to fit within achievable LTE speeds. 




\bibliography{references.bib}

\end{document}